\documentclass[runningheads]{llncs}

\usepackage{tikz}
\usetikzlibrary{shapes.geometric, arrows.meta, positioning, calc}
\usepackage{graphicx}
\usepackage[T1]{fontenc}
\usepackage{booktabs}
\usepackage{microtype}
\usepackage{array}
\usepackage{stfloats}
\usepackage{algorithm}
\usepackage{algpseudocode}
\usepackage{amsmath}
\usepackage{url}
\usepackage{float}

\usepackage{marvosym}
\begin{document}
\title{Benchmarking Foundation Models for Bone Scaffold Pore Detection: A Comparative Analysis of Few-Shot Performance}
\titlerunning{Benchmarking Foundation Models for Bone Scaffold Pore Detection}
\author{Tri Tin Vo\inst{1}\textsuperscript{(\Letter)}\orcidID{0009-0003-8007-3663} \and
Vu Thien An Ngo\inst{1}\orcidID{0009-0001-4805-381X} \and
Nhi Kieu\inst{1}\orcidID{0009-0002-8046-2311} \and
Cuong Hung Luu\inst{2}\orcidID{0000-0002-0919-5911} \and
Giang V.H. Phan\inst{3}\orcidID{0000-0002-0866-8131} \and
Kien Nguyen\inst{1}\orcidID{0000-0002-3466-9218} \and
Clinton Fookes\inst{1}\orcidID{0000-0002-8515-6324}}

\authorrunning{T. T. Vo et al.}

\institute{Queensland University of Technology, Brisbane QLD 4000, Australia\\
\email{\{n12501034, ngov2, v.kieu, k.nguyenthanh, c.fookes\}@qut.edu.au} \and
Griffith University, Nathan QLD 4111, Australia\\
\email{cuong.luu@griffithuni.edu.au} \and
Ton Duc Thang University, Ho Chi Minh City, Vietnam\\
\email{phanvuhoanggiang@tdtu.edu.vn}}
%
\maketitle              
\begin{abstract}
Evaluating the mechanical properties of bone scaffolds conventionally requires destructive testing or costly Finite Element Analysis, while deep learning alternatives are constrained by the scarcity of labeled biomaterial imaging data. This study introduces the first expert-annotated, pixel-dense dataset of biopolymer bone-scaffold SEM images, spanning five magnification series, and uses it to benchmark two state-of-the-art pre-trained foundation models under a 12-shot fine-tuning regime. SAM3, an instance segmenter, and YOLOv26, a bounding-box detector, feed the predicted pore geometry to regression heads that estimate Young's modulus and compressive yield strength. Our central finding is that both models fail by pore density rather than magnification: YOLOv26's predicted pore counts saturate and porosity estimation collapses on the most crowded frames, whereas SAM3 over-fragments sparse pore fields and merges adjacent pores in dense ones, producing eight catastrophic failures against three for YOLOv26. Averaged across the 44-image evaluation set, YOLOv26 achieves lower porosity error (19.9\% vs.\ 23.1\% MAE), higher precision, and a higher F1 score, whereas SAM3 offers pixel-level instance segmentation that YOLOv26's bounding boxes cannot reproduce. We conclude that model selection should be driven by pore density rather than SEM magnification alone.

\end{abstract}
\section{Introduction}
A bone scaffold's mechanical performance is, in practice, established through destructive characterization. Standard validation combines physical compression testing with Finite Element Analysis (FEA), and both routes are slow: FEA requires an accurate mesh reconstruction of the scaffold's internal pore architecture, while physical testing irreversibly consumes the sample. For biopolymer scaffolds fabricated from alginate, chitosan, gelatin, and silk fibroin, this bottleneck is compounded by batch-to-batch variability in pore geometry, even under fixed processing parameters. Consequently, every formulation change necessitates a new testing cycle before mechanical suitability can be confirmed.

Deep learning has been proposed as a faster substitute. Convolutional networks have been applied to analyze 3D scaffold microarchitecture directly from micro-CT and SEM grids \cite{abedini2026,salerno2026}. Graph neural networks model the irregular pore-and-strut connectivity of trabecular bone to predict load paths and stress concentration \cite{abedini2026}, while U-Net variants estimate local micro-strain in TPMS lattices without performing a full micro-FE solve \cite{wu2023}. Semantic segmentation has been used in a similar manner to track bone ingrowth into biodegradable scaffolds from micro-computed tomography (micro-CT) volumes \cite{ganeshaaraj2022}. Karaca and Dikici trained deep networks directly on Scanning Electron Microscopy (SEM) images to size pores and windows in tissue-engineering scaffolds, using the same imaging modality and target geometry as the present study, though their dataset was not constrained to a few-shot regime \cite{karaca2024}.

None of this work yet escapes the data requirement. Every model cited above was trained on a dataset large enough to support standard supervised learning, a luxury rarely afforded in bone and biomaterial imaging. Micro-CT and SEM acquisition is expensive, scaffold geometries are batch-specific, and the manual annotation required to produce training masks is slow and inconsistent between annotators \cite{salerno2026}. A laboratory developing a new scaffold formulation will often have a dozen usable images rather than a thousand. This mismatch between the requirements of deep learning and realistic ouput of biomaterials research is the practical motivation for adopting few-shot learning here, rather than an academic preference for smaller training sets. Few-shot techniques have already demonstrated their capacity to circumvent this constraint in cardiac, abdominal, and pulmonary imaging \cite{pachetti2023}, yet their application to scaffold microstructure remains largely untested.

The main contributions of this paper are as follows:
\begin{itemize}
    \item \textbf{Expert-annotated Dataset.} A selective 56-image dataset of SEM scaffold images, fabricated from natural polymers via freeze-drying, was annotated by domain experts to ensure reliability, as the resulting porous architectures are inherently more complex to delineate than those produced by emulsion templating \cite{karaca2024}.

    \item \textbf{Few-shot Benchmarks on Scaffold SEM Imagery.} Two state-of-the-art foundation models, the Segment Anything Model (SAM3) and YOLOv26 models, are assessed a 12-shot fine-tuning on SEM images for pore localization in biopolymer bone scaffolds.

    \item \textbf{Non-destructive Mechanical-property Prediction.} The detected pore features feed regression heads that predicts Young's modulus \cite{boccaccini1993} and compressive yield strength, aiming to eliminate the need for physical or destructive testing.

    \item \textbf{Diagnostic Comparison.} We document where each model succeeds and fails, showing that SAM3 and YOLOv26 trade spatial precision against inference robustness in a scale-dependent way, thereby providing practitioners with the guidance needed to select a model before deploying it on unlabeled scaffold imagery.
\end{itemize}

The remainder of this paper is structured as follows. Section 2 surveys related works in this field. Section 3 details the proposed methodology, showcasing the pipeline design and the foundational models utilized. Section 4 outlines the experimental setup, including the dataset description and evaluation metrics. Section 5 presents the comparative analysis of model performance, physical metrics, and failure cases. Section 6 summarizes the key findings, addresses their implications for resolving data limitations, and outlines directions for future work.

\section{Related Works}

Traditionally, the morphological characterization and performance evaluation of these microarchitectures have relied heavily on Computer-Aided Design (CAD) and computationally expensive FEA \cite{abedini2026}. Recently, Artificial Intelligence (AI) and Deep Learning (DL) have emerged to automate the detection, segmentation, and evaluation of scaffold microstructures, offering a paradigm shift away from conventional trial-and-error modeling \cite{wu2023}. 

Convolutional Neural Networks (CNNs) have proven to be highly effective in analyzing intricate 3D microarchitectures of bone scaffolds because they are specifically tailored to handle grid data, such as 2D/3D images and micro-CT scans \cite{abedini2026,salerno2026}. Beyond grid-based CNNs, U-Net models have been deployed to predict locally varying micro-strain components in Triply Periodic Minimal Surface (TPMS) unit cells, bypassing computationally prohibitive micro-FE analyzes \cite{wu2023}. Graph Neural Networks (GNNs) have additionally been employed to model the irregular, non-Euclidean connectivity of trabecular bone and complex scaffold pore networks. By representing individual pores or struts as nodes, GNNs capture structural hierarchies to improve prediction of load distribution, stress concentrations, and fluid pathways \cite{abedini2026}. 

Despite the success of CNNs and GNNs, a major gap persists in the widespread application of deep learning to scaffold design: the severe reliance on data. Deep learning models typically require large-scale labeled datasets to achieve high predictive performance \cite{pachetti2023}. However, acquiring large, high-quality imaging datasets (e.g., SEM or phase-contrast micro-CT) for highly specific scaffold architectures is notably difficult, expensive, and computationally burdensome. Furthermore, naturally derived and bio-inspired scaffolds often exhibit irregular, anisotropic internal architectures that complicate morphological characterization. Manually annotation of these complex microstructures to create reliable training masks is extremely labor-intensive and highly susceptible to inter-operator variability and user bias \cite{salerno2026}. Collectively, these fundamental limitations create a significant bottleneck that impedes the rapid scaling and clinical translation of AI-driven scaffold evaluation.

To address the critical issue of data scarcity in medical image analysis, researchers have begun exploring Few-Shot Learning (FSL) techniques. FSL significantly reduces dependence on massive datasets by leveraging meta-learning paradigms, enabling models to adapt to new, unseen tasks using only a small number of labeled samples. Although FSL, along with semi-supervised learning strategies, has demonstrated success in broader medical imaging domains such as cardiac, abdominal, and pulmonary disease classification \cite{pachetti2023}, its application to the microstructural segmentation and mechanical characterization of bone scaffolds remains largely untapped.

Overall, current deep learning methodologies for bone scaffold optimization are hindered by the immense difficulty of obtaining high-quality micro-CT datasets and the labor-intensive, bias-prone nature of manual microstructural annotation. To overcome these challenges, this study proposes a novel few-shot learning framework for the mechanical property prediction of bone scaffolds. By leveraging advanced foundational vision models, specifically the Segment Anything Model (SAM3) and YOLOv26, the proposed FSL approach efficiently identifies and extracts key structural features, such as localizing pores via bounding boxes, using only a minimal number of annotated training examples. Within our pipeline, these accurately localized morphological features are subsequently mapped to physical mechanical properties using a targeted regression head. This methodology provides a rapid, non-destructive, and computationally efficient alternative to conventional physical testing and FEA, effectively resolving the data constraints that limit conventional deep learning models.

\section{Methodology}
This study benchmarks state-of-the-art foundation models on SEM scaffold images. Through a few-shot learning pipeline, the morphological features are mapped to their corresponding mechanical properties. The two selected foundation models are: (1) the Segment Anything Model SAM3 \cite{sam3}, a contour-precise instance segmenter, and (2) YOLOv26 \cite{yolov26}, a rapid bounding-box detector. Evaluating these architectures under identical, strictly data-limited conditions enables characterization of their respective strengths and failure modes in the biomaterials context. The remainder of this section describes the few-shot fine-tuning framework established in this work. (Fig.~\ref{fig:pipeline_workflow}).

\subsection{Pipeline Design}

\begin{figure}[htbp]
    \centering
    \resizebox{\textwidth}{!}{
    \begin{tikzpicture}[
        node distance=1.2cm and 2.2cm,
        font=\sffamily\small,
        imgbox/.style={draw, gray!50, thick, inner sep=0.5pt, line width=1pt},
        box/.style={draw, rectangle, rounded corners, minimum height=1.0cm, align=center, thick},
        step1/.style={box, fill=cyan!10, draw=cyan!80!black, minimum width=4.2cm},
        step2/.style={box, fill=teal!10, draw=teal!80!black, minimum width=4.2cm},
        step3/.style={box, fill=purple!10, draw=purple!80!black, minimum width=3.8cm},
        subbox/.style={draw, rectangle, rounded corners=2pt, fill=white, draw=gray, minimum width=3.4cm, minimum height=1.4cm, align=center, font=\sffamily\bfseries\footnotesize},
        labelnode/.style={align=center, font=\sffamily\bfseries\footnotesize, gray!80!black},
        arrow/.style={-{Stealth[scale=0.8]}, thick, draw=gray!75}
    ]

        \node (raw_img) [imgbox] {\includegraphics[width=4.4cm, height=4.4cm]{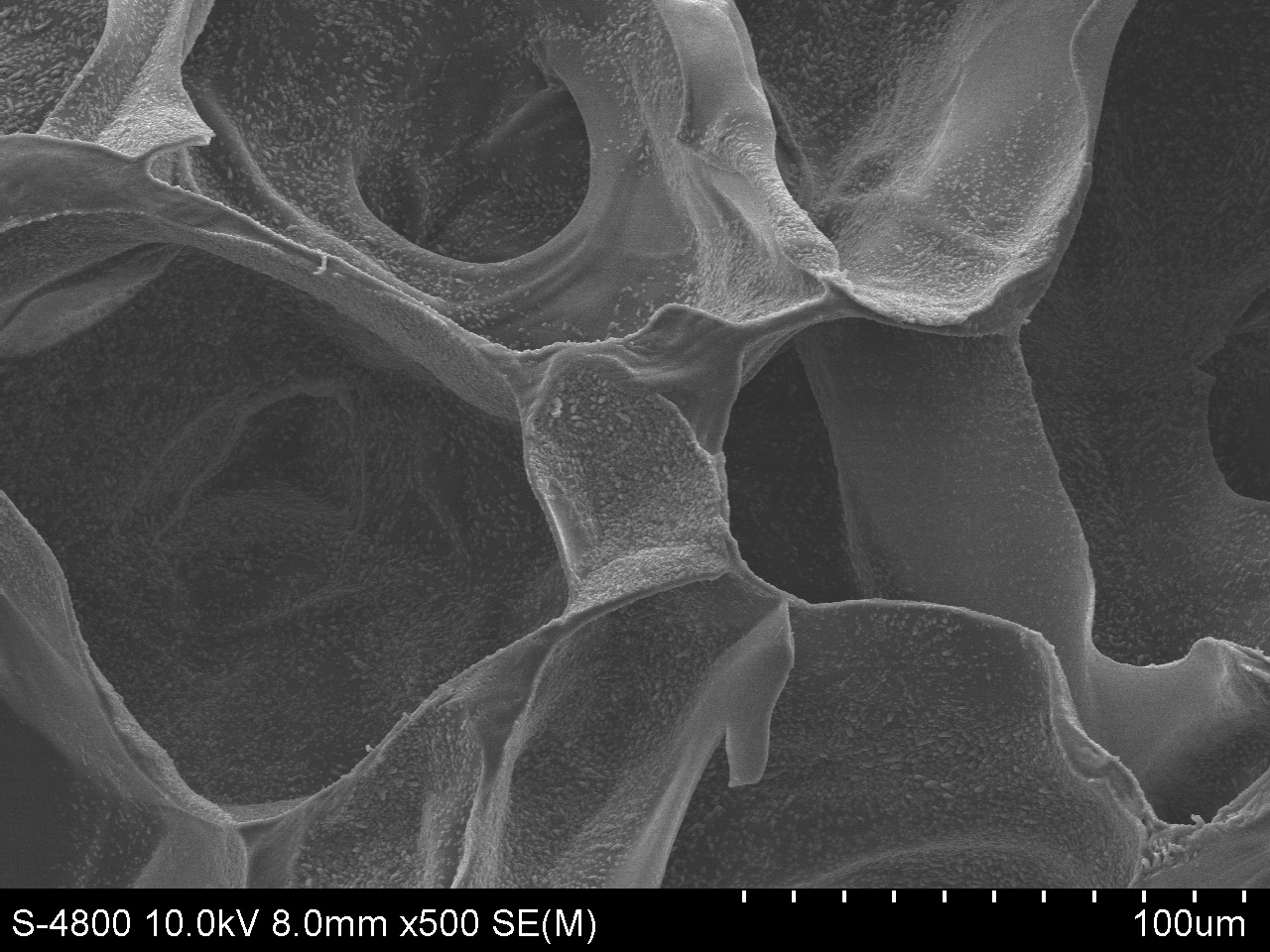}};
        \node (raw_lbl) [labelnode, above=0.1cm of raw_img] {1. Raw SEM Image};

        \node (clean_img) [imgbox, below=1.4cm of raw_img] {\includegraphics[width=4.4cm, height=4.4cm]{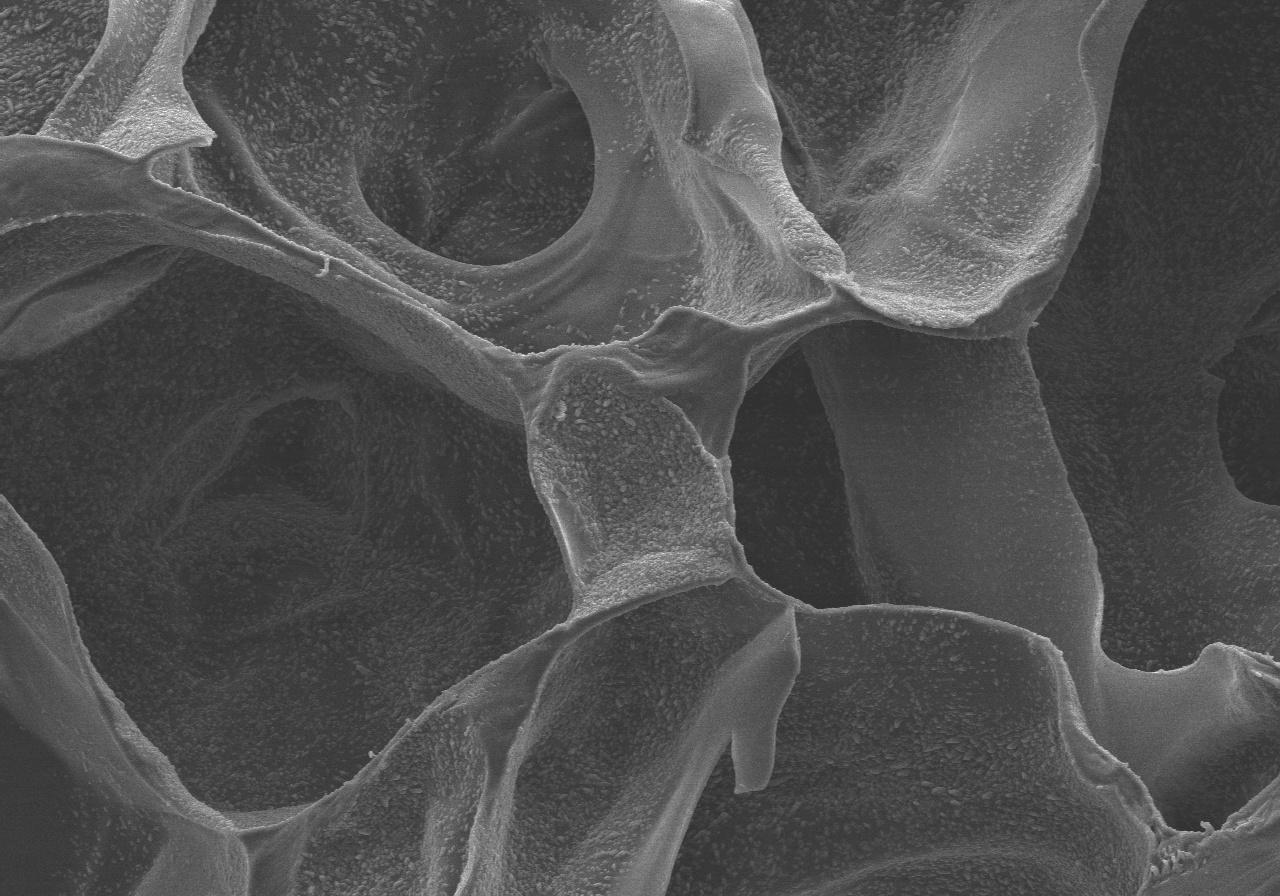}};
        \node (clean_lbl) [labelnode, below=0.1cm of clean_img] {2. Cleaned Image};

        \node (models) [step3, minimum height=4.8cm] at ($(clean_img.center)!0.5!(raw_img.center) + (5.5, 0)$) {};
        \node (models_lbl) [labelnode, above=0.1cm of models] {3. Few-Shot Models};
        
        \node (yolo) [subbox, below=10pt of models.north] {YOLOv26s\\
            \normalfont\scriptsize $\bullet$ Bounding Box Regressor\\
            \normalfont\scriptsize $\bullet$ Anchor-free CNN\\
            \normalfont\scriptsize $\bullet$ Finetuned Head};
            
        \node (sam) [subbox, below=12pt of yolo] {SAM3\\
            \normalfont\scriptsize $\bullet$ Instance Segmenter\\
            \normalfont\scriptsize $\bullet$ Few-shot Text Prompt\\
            \normalfont\scriptsize $\bullet$ ViT Image Encoder};

        \node (yolo_out) [imgbox] at (raw_img.center -| {$(models.east) + (3.6, 0)$}) {\includegraphics[width=4.4cm, height=4.4cm]{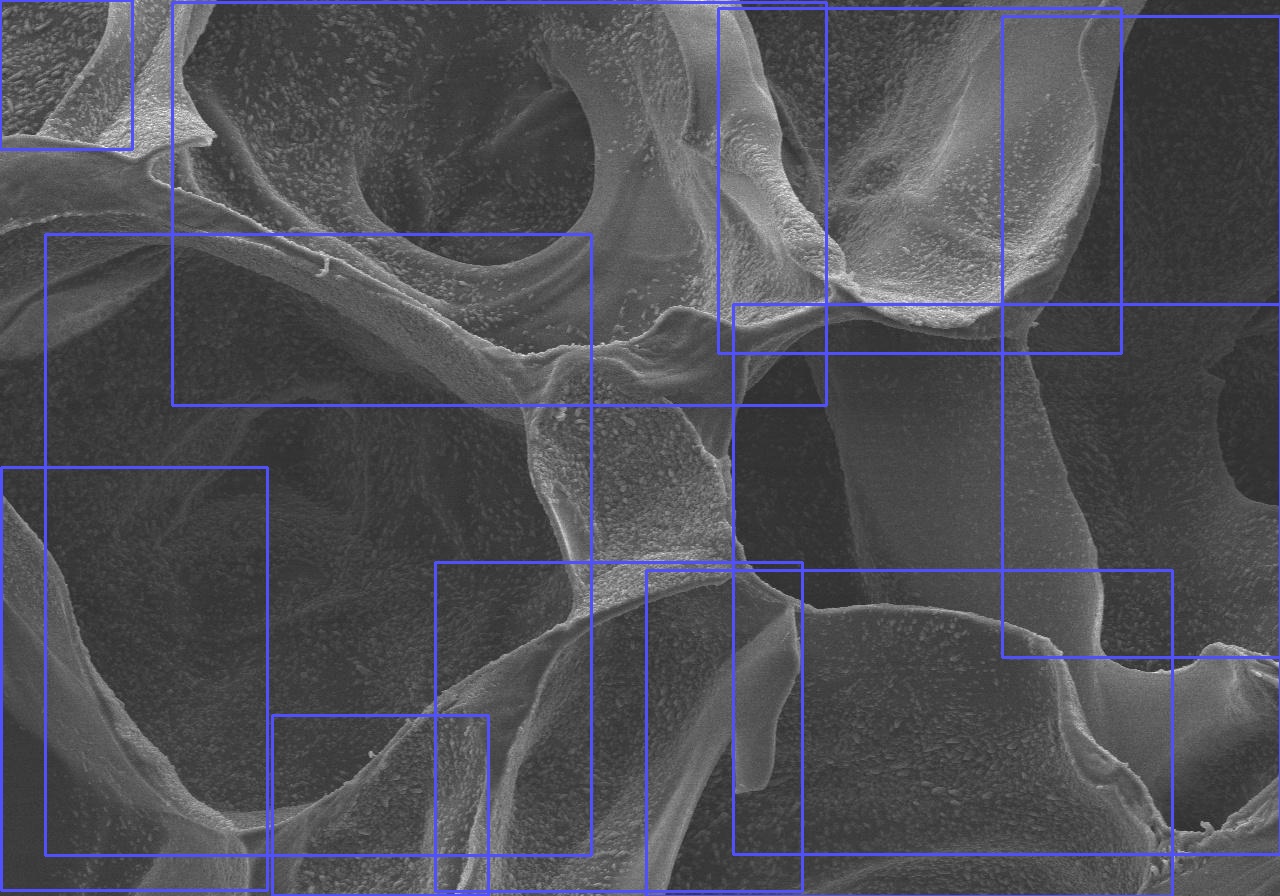}};
        \node (yolo_lbl) [labelnode, above=0.1cm of yolo_out] {YOLO Detections};

        \node (sam_out) [imgbox] at (clean_img.center -| yolo_out.center) {\includegraphics[width=4.4cm, height=4.4cm]{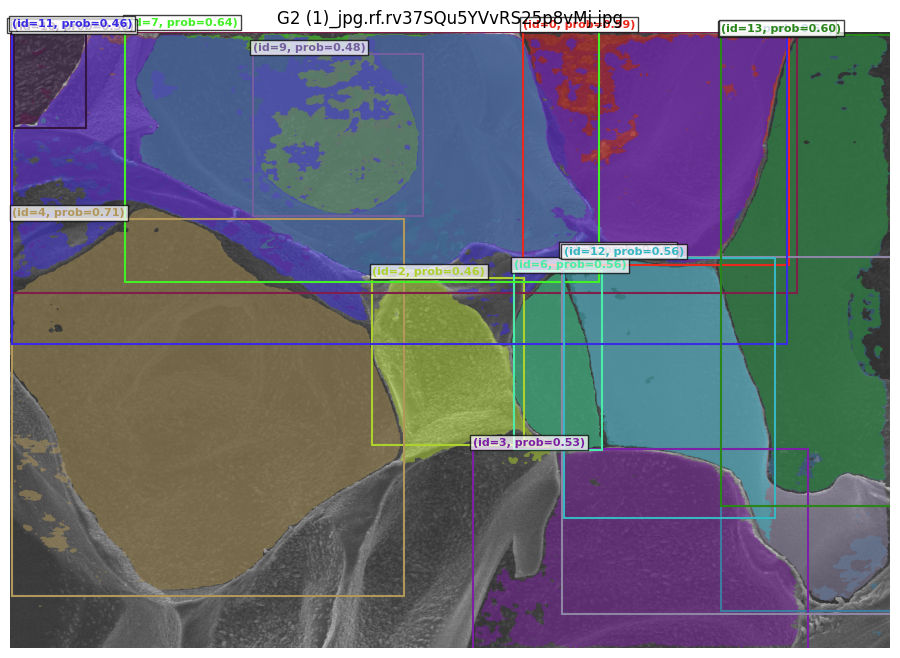}};
        \node (sam_lbl) [labelnode, below=0.1cm of sam_out] {SAM3 Segmentations};

        \draw [arrow] (raw_img.south) -- (clean_img.north);
        
        \draw [arrow] (clean_img.east) -| (models.south);
        
        \draw [arrow] (yolo.east) -- (yolo_out.west);
        \draw [arrow] (sam.east) -- (sam_out.west);

    \end{tikzpicture}
    }
    \caption{The few-shot learning scaffold analysis pipeline workflow. (1) Raw SEM image input; (2) Image denoising and cropping; (3) Few-shot detection models (YOLOv26 and SAM3); (4) Model Predictions.}
    \label{fig:pipeline_workflow}
\end{figure}

The pipeline consists of three sequential stages in Fig.~\ref{fig:pipeline_workflow}. First, raw SEM images undergo noise reduction to minimize the artifacts and are cropped to the region of interest through black-strip removal, then resized to a uniform resolution of $1280 \times 896$. Second, the two foundation models, SAM3 and YOLOv26, undergo few-shot fine-tuning processes to localize scaffold pores. YOLOv26 produces bounding boxes, while SAM3 additionally resolves pixel-level instance segmentation masks. This stage aims to isolate the morphological features of the bone scaffolds. Third, the pipeline estimates scaffold mechanical properties based from the predicted features. For each image, a morphological feature vector $\phi$ is assembled from the predicted regions, comprising the width, height, and area of each detected pore together with the aggregate pore density and porosity. This feature vector is mapped to the target mechanical properties, Young's modulus and compressive yield strength, by a regression head $g_\theta : \phi \rightarrow \hat{y}$. The YOLOv26 features are mapped through a Random Forest regressor, whereas the SAM3 features are mapped through the Gibson-Ashby physical model \cite{gibson1982}, as reflected in the mechanical correlation results (Fig.~\ref{fig:mechanical_correlation}). We note that this design couples each detector to a specific regressor, so the mechanical-property results in Fig.~\ref{fig:mechanical_correlation} reflect detector-and-regressor pipelines rather than the detectors in isolation. This is intentional: it mirrors how each model would realistically be deployed. A fully factorial comparison, crossing both detectors with both regressors, is left to future work.

\subsection{Few-shot Learning Framework and Models}
To address the inherent scarcity and high annotation cost of biomaterial scaffold SEM data, this study formulates the task as a FSL problem with $K$-shot. FSL reduces the dependence on large annotated datasets by adapting pre-trained representations to a new domain from only a small number of labeled examples, a paradigm that has proven effective in broader medical imaging settings. To assess the performance of the selected foundation models on these biopolymer materials, $12$-shot fine-tuning was applied ($K=12$). The two state-of-the-art foundation models are as follows:

\begin{itemize}
    \item \textbf{SAM3 \cite{sam3}}: a powerful Vision Transformer (ViT)-based instance segmenter. SAM3 was utilized for its text-prompting and generalization capabilities in FSL, enabling precise segmentation of complex pore boundaries in scaffold SEM images. The backbone was pre-trained on an extensive dataset of 4M distinct text labels and 52M instance segmentation masks.
    \item \textbf{YOLOv26 \cite{yolov26}}: an anchor-free convolutional bounding-box regressor in widespread industrial use. It was employed as a rapid object-detection model to localize scaffold pores directly under strict few-shot constraints. The backbone was pre-trained on 756K images with 11.6M bounding boxes.
\end{itemize}

\begin{algorithm}[htbp]
\caption{SEM-Based Scaffold Analysis Pipeline}
\label{alg:pipeline}
\scriptsize
\begin{algorithmic}[1]

\Require Raw SEM images $\mathcal{D} = \{I_1, I_2, \ldots, I_n\}$
\Ensure Predicted mechanical properties $\hat{y}$

\vspace{4pt}
\State \textbf{// Step 1: Data Preprocessing}
\For{each image $I_i \in \mathcal{D}$}
    \State $I_i \leftarrow \text{Denoise}(I_i)$ \Comment{reduce electron beam artifacts}
    \State $I_i \leftarrow \text{Resize}(I_i,\ 1280 \times 896)$
    \State $\mathcal{D} \leftarrow \mathcal{D} \cup \text{Augment}(I_i)$ \Comment{flip, rotate, brightness}
\EndFor
\vspace{4pt}
\State \textbf{// Step 2: Pre-Trained Object Detection}
\State Load $f_{\text{SAM3}}$,\ $f_{\text{YOLOv26}}$ \Comment{foundation models}
\For{each image $I_i \in \mathcal{D}$}
    \State $\mathcal{B}^{\text{SAM3}}  \leftarrow f_{\text{SAM3}}(I_i)$
    \State $\mathcal{B}^{\text{YOLO}} \leftarrow f_{\text{YOLOv26}}(I_i)$
    \State $\mathcal{B}_i \leftarrow \mathcal{B}^{\text{SAM3}} \cup \mathcal{B}^{\text{YOLO}}$
    \For{each box $b \in \mathcal{B}_i$}
        \State $\phi_i \mathrel{+}= \left[\,w_b,\ h_b,\ a_b,\ \rho_{\text{pore}},\ \phi_{\text{porosity}}\,\right]$
    \EndFor
\EndFor

\vspace{4pt}
\State \textbf{// Step 3: Mechanical Property Prediction}
\State Initialize regression head $g_\theta : \phi \rightarrow \hat{y}$
\For{each feature vector $\phi_i$}
    \State $\hat{y}_i \leftarrow g_\theta(\phi_i)$ \Comment{e.g.\ Young's modulus, compressive strength}
\EndFor

\State \Return $\{\hat{y}_i\}_{i=1}^{n}$

\end{algorithmic}
\end{algorithm}

The two models therefore trade spatial precision against computational cost and inference speed: SAM3 resolves contour-hugging masks, while YOLOv26 offers fast, rectangular approximations. To simulate realistic low-annotation scenarios, both models were fine-tuned using 12 images drawn from the 56-image dataset via stratified random sampling, preserving the distribution of pore density across the support set. The remaining 44 images were held out exclusively for evaluation.

Given the disparate scales of the two models, we adopted a different parameter-efficient fine-tuning strategy for each. For SAM3, the heavy ViT image encoder and the segmentation head were frozen throughout training. Only the text-vision alignment layer and the final two prediction layers were set as trainable, yielding approximately 1.5\% trainable parameters (12.7M out of 848M). This design preserves the generalized visual representations learned by the foundation model while allowing its output distribution to the scaffold pore domain. For YOLOv26, the convolutional backbone (layers 0-9) was largely frozen, with the remaining 14 neck and detection head layers (layers 10-23) set as trainable, corresponding to 58.3\% of the total 10.01M parameters (5.84M learnable parameters). Both models were fine-tuned with early stopping, using a patience of 20 epochs on validation loss.

Algorithm \ref{alg:pipeline} summarizes the end-to-end procedure. Raw images are first cleaned; the fine-tuned SAM3 and YOLOv26 models then produce pore detections, from which the morphological feature vector $\phi$ is assembled. Finally, the regression head maps these features to the predicted mechanical properties.

\begin{figure}[htbp]
\centering
\setlength{\tabcolsep}{1pt}
\begin{tabular}{ccc}
  \includegraphics[width=0.24\textwidth]{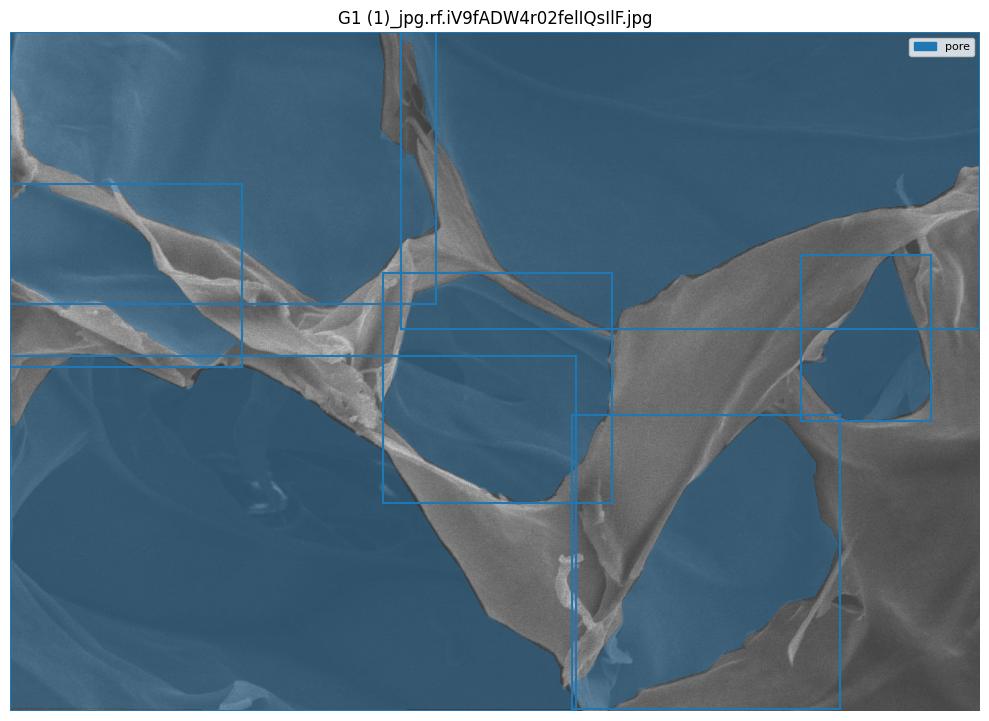} &
  \includegraphics[width=0.24\textwidth]{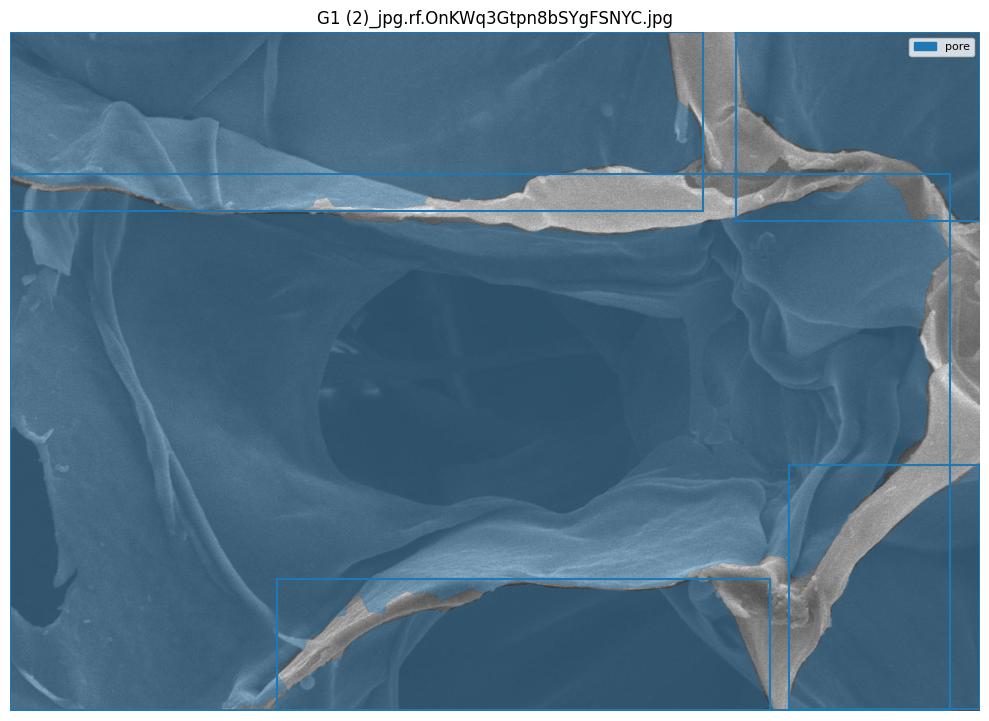} &
  \includegraphics[width=0.24\textwidth]{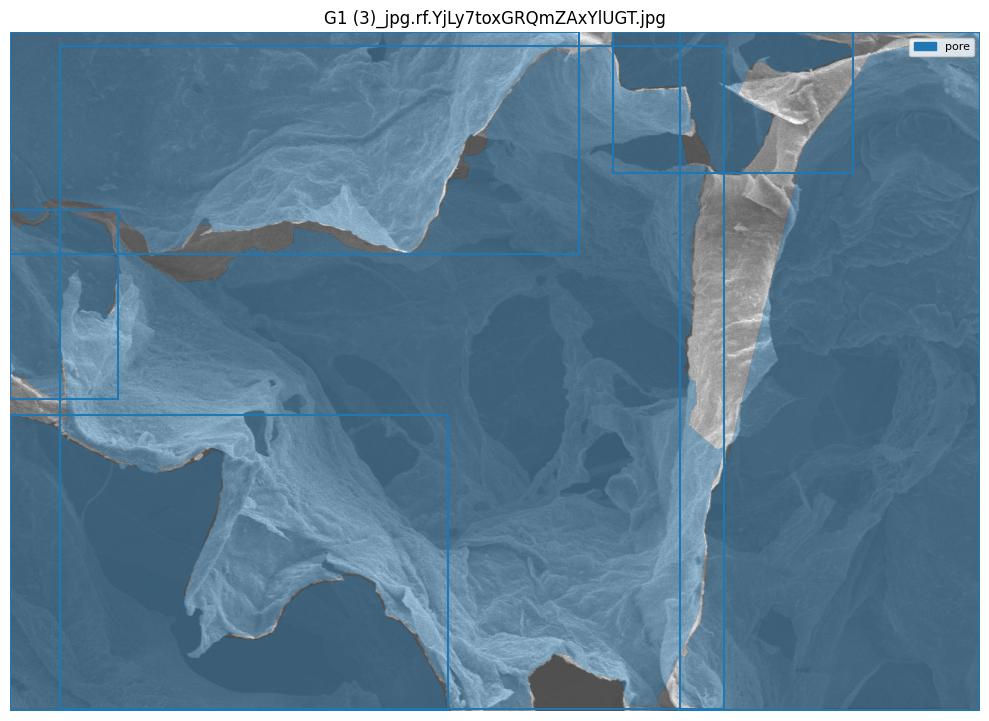} \\
  (a) G1 - Sample 1 & (b) G1 - Sample 2 & (c) G1 - Sample 3 \\[3pt]

  \includegraphics[width=0.24\textwidth]{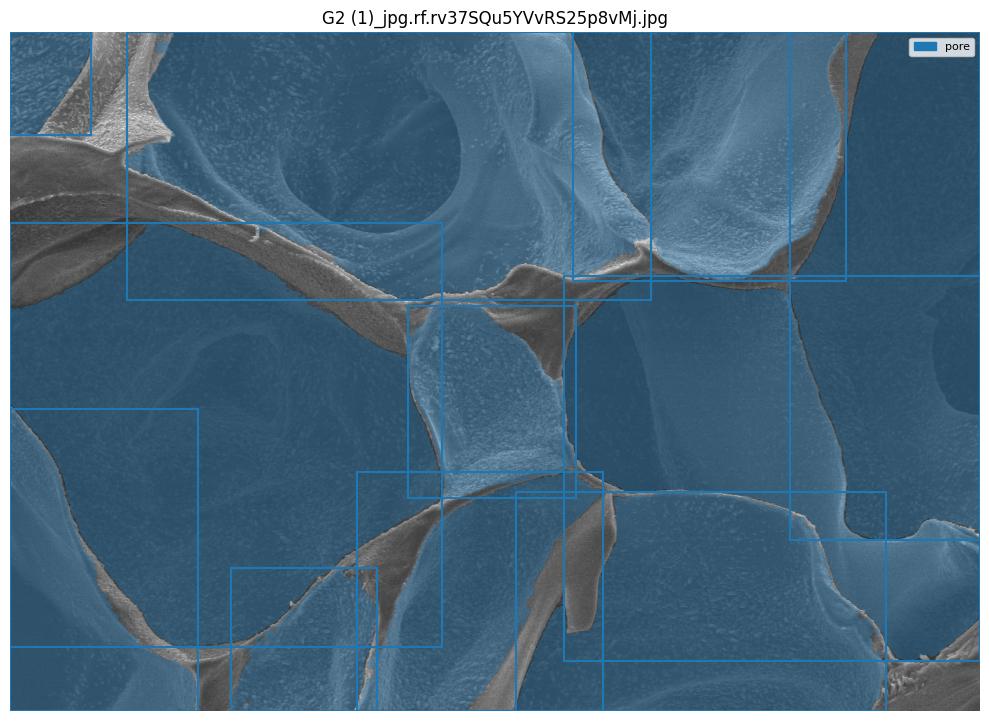} &
  \includegraphics[width=0.24\textwidth]{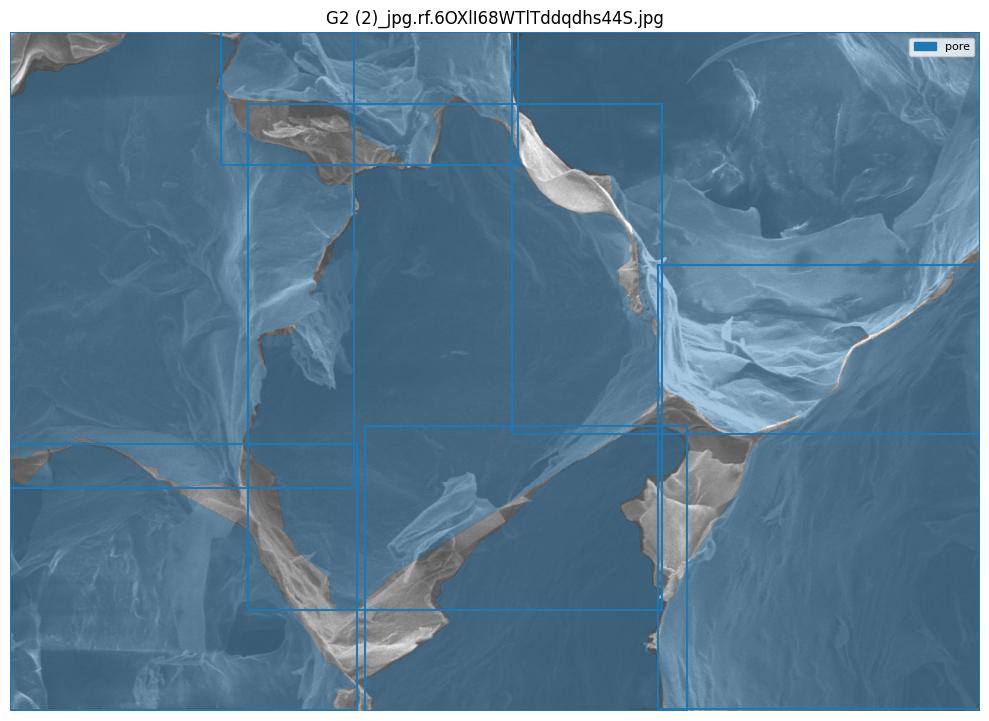} &
  \includegraphics[width=0.24\textwidth]{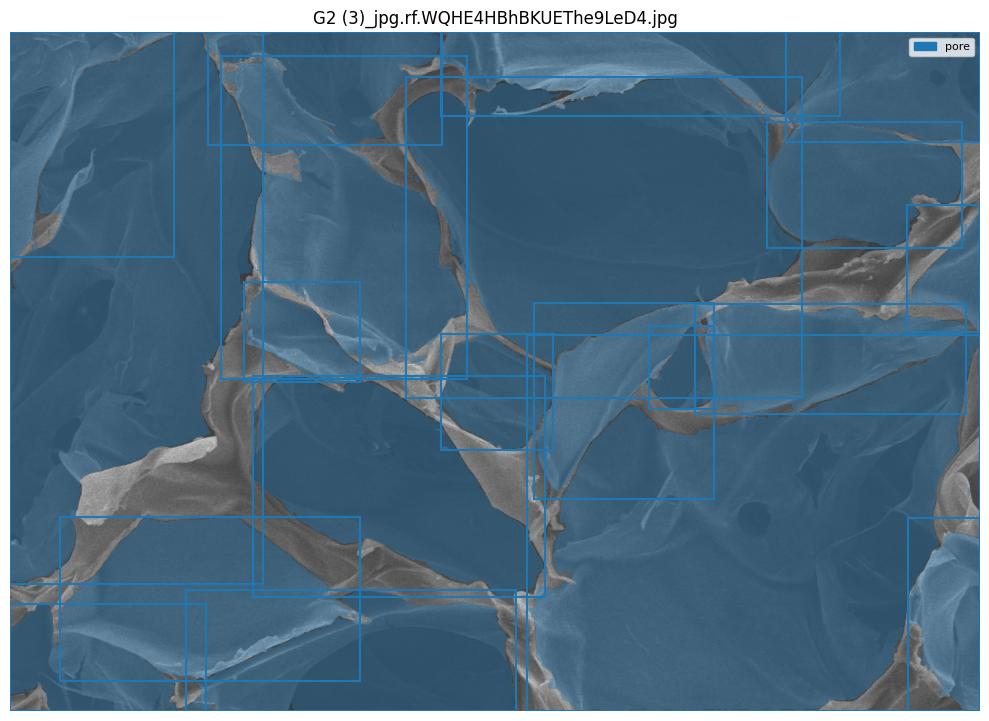} \\
  (d) G2 - Sample 1 & (e) G2 - Sample 2 & (f) G2 - Sample 3 \\[3pt]

  \includegraphics[width=0.24\textwidth]{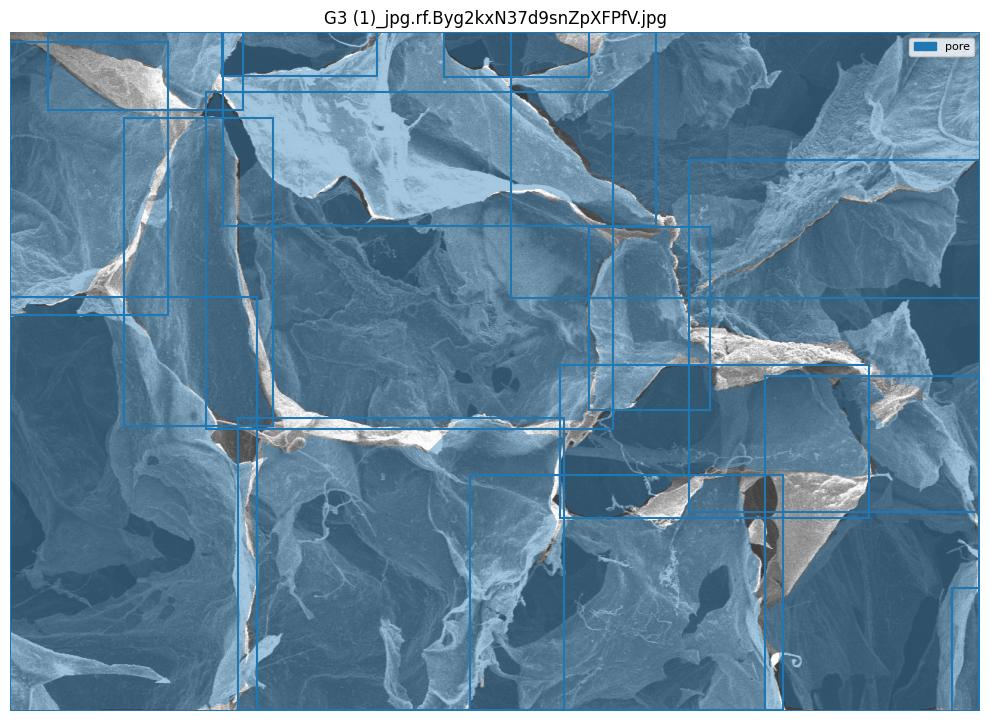} &
  \includegraphics[width=0.24\textwidth]{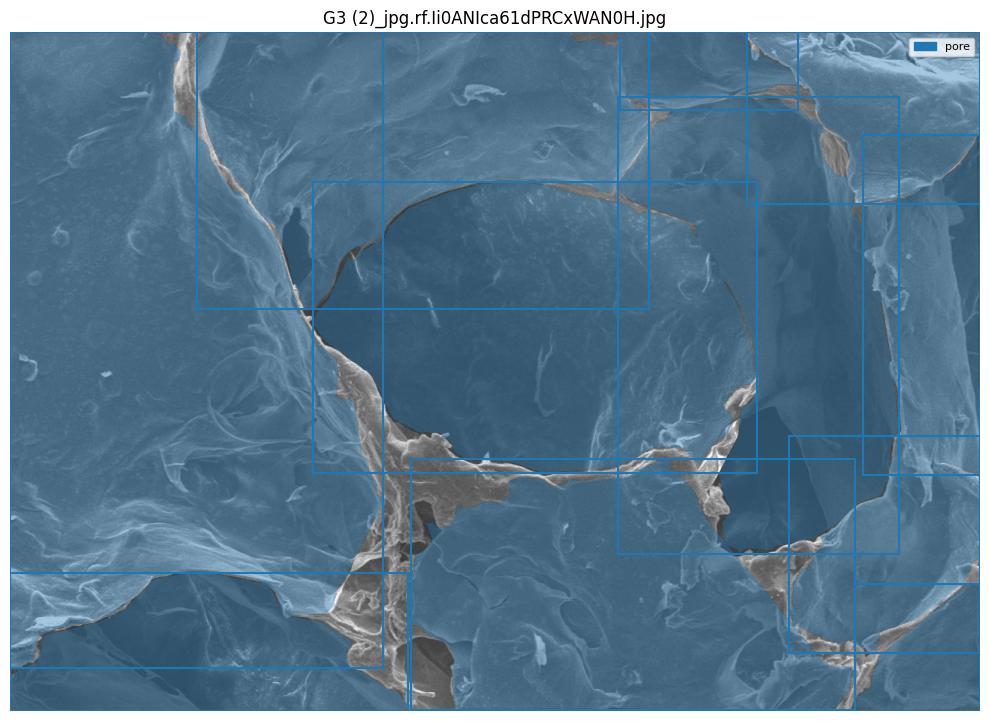} &
  \includegraphics[width=0.24\textwidth]{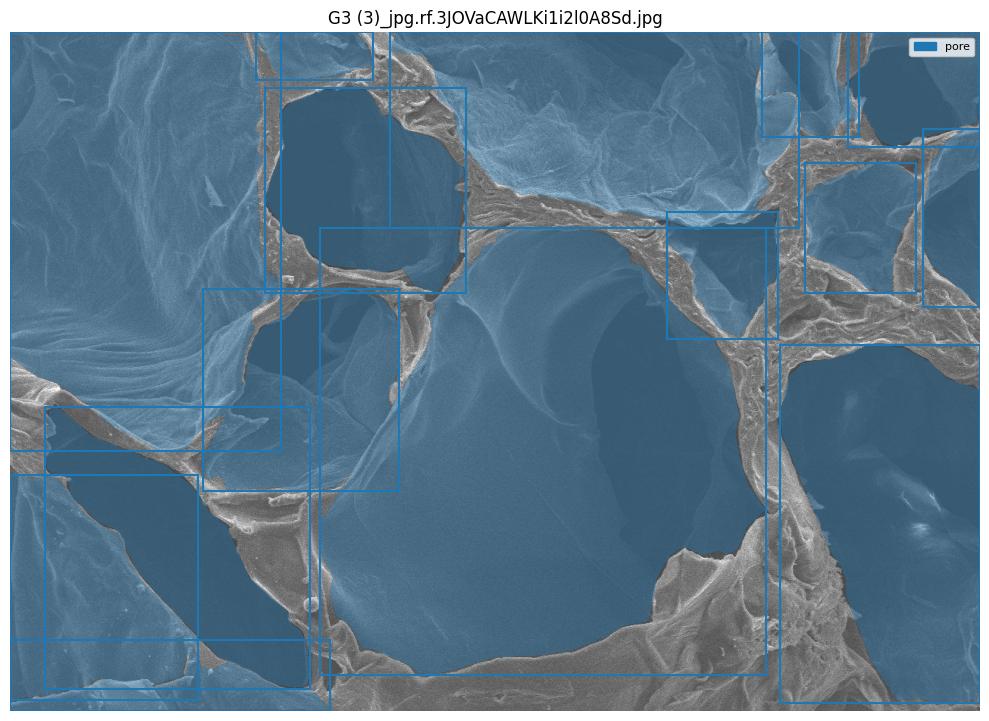} \\
  (g) G3 - Sample 1 & (h) G3 - Sample 2 & (i) G3 - Sample 3 \\[3pt]

  \includegraphics[width=0.24\textwidth]{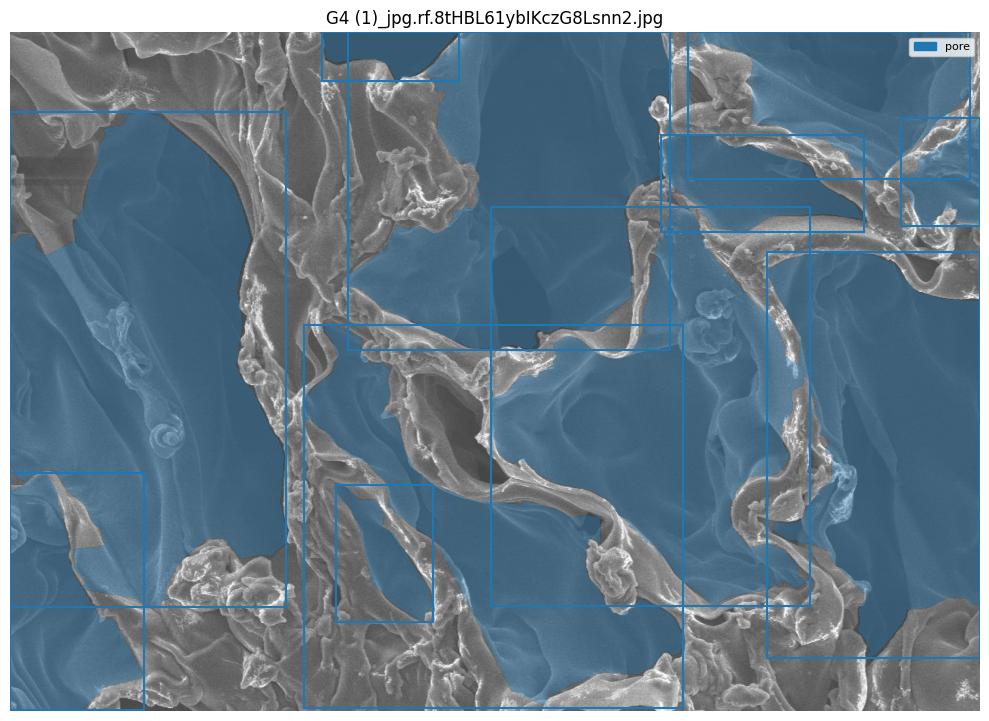} &
  \includegraphics[width=0.24\textwidth]{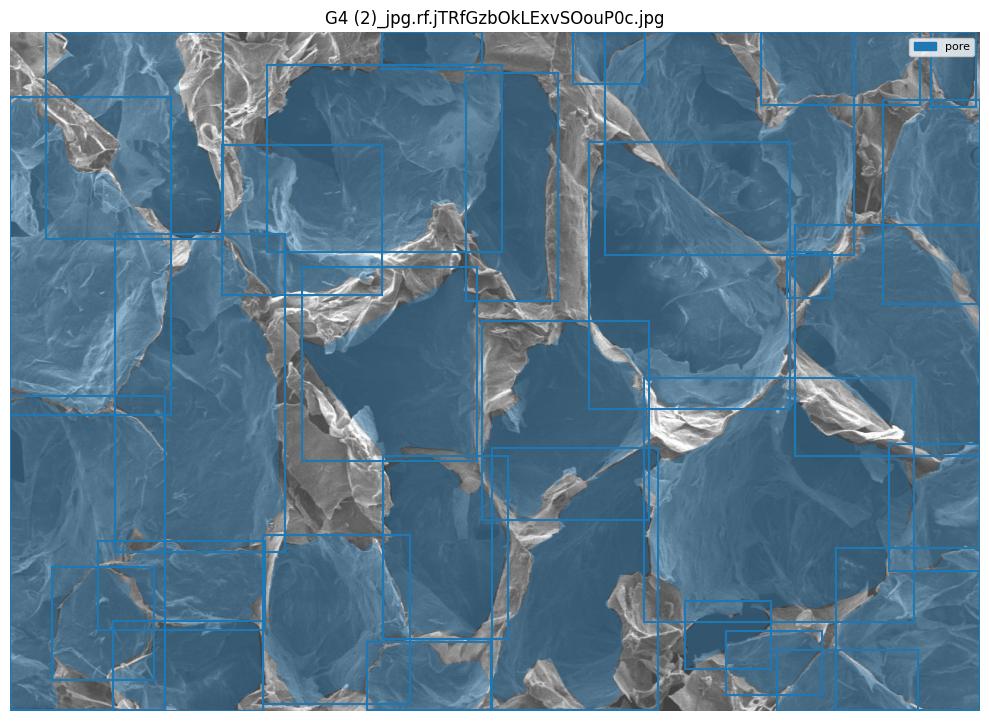} &
  \includegraphics[width=0.24\textwidth]{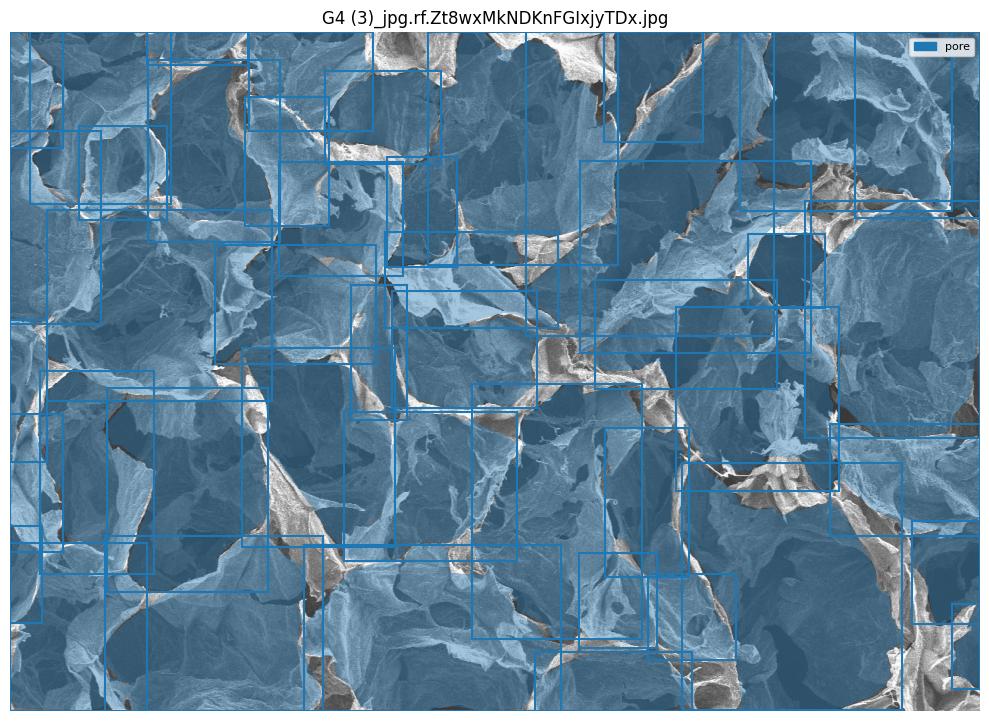} \\
  (j) G4 - Sample 1 & (k) G4 - Sample 2 & (l) G4 - Sample 3 \\[3pt]

  \includegraphics[width=0.24\textwidth]{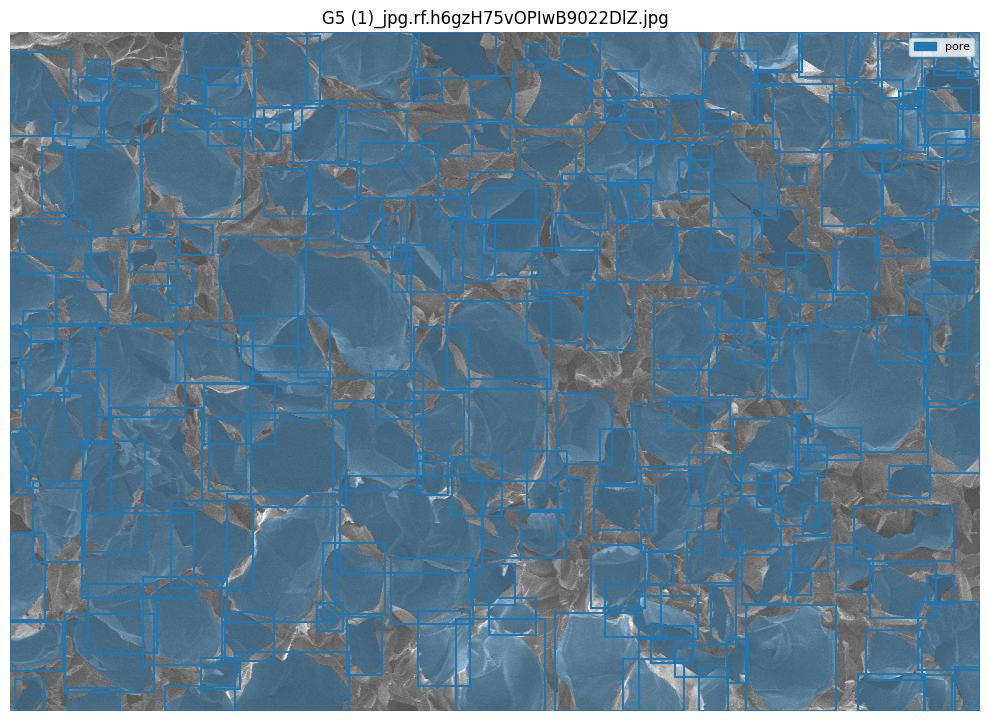} &
  \includegraphics[width=0.24\textwidth]{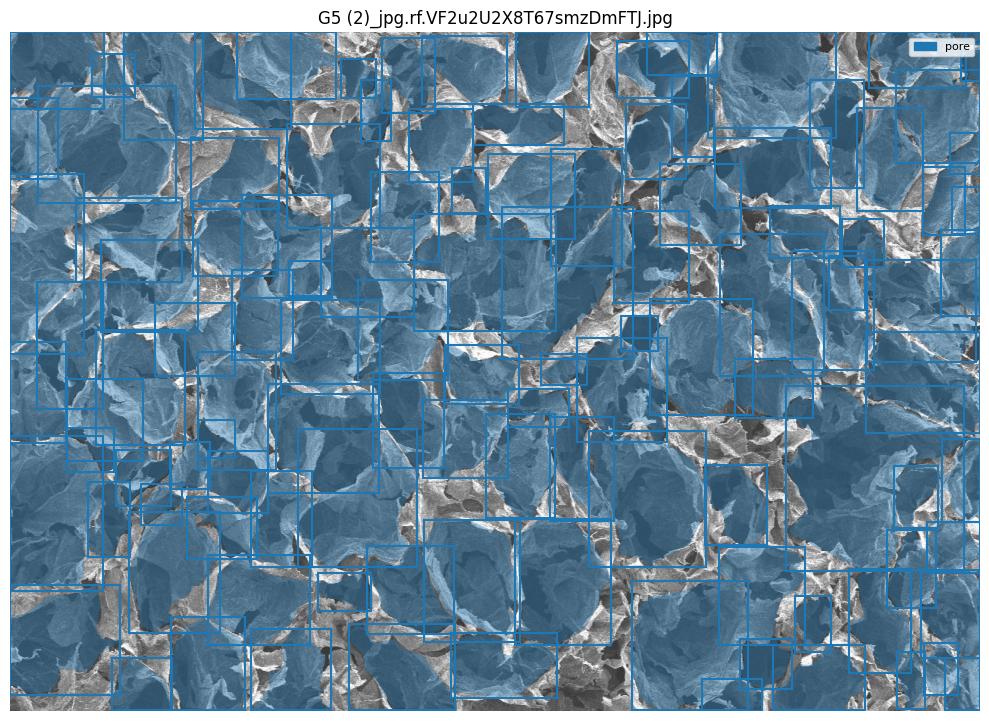} &
  \includegraphics[width=0.24\textwidth]{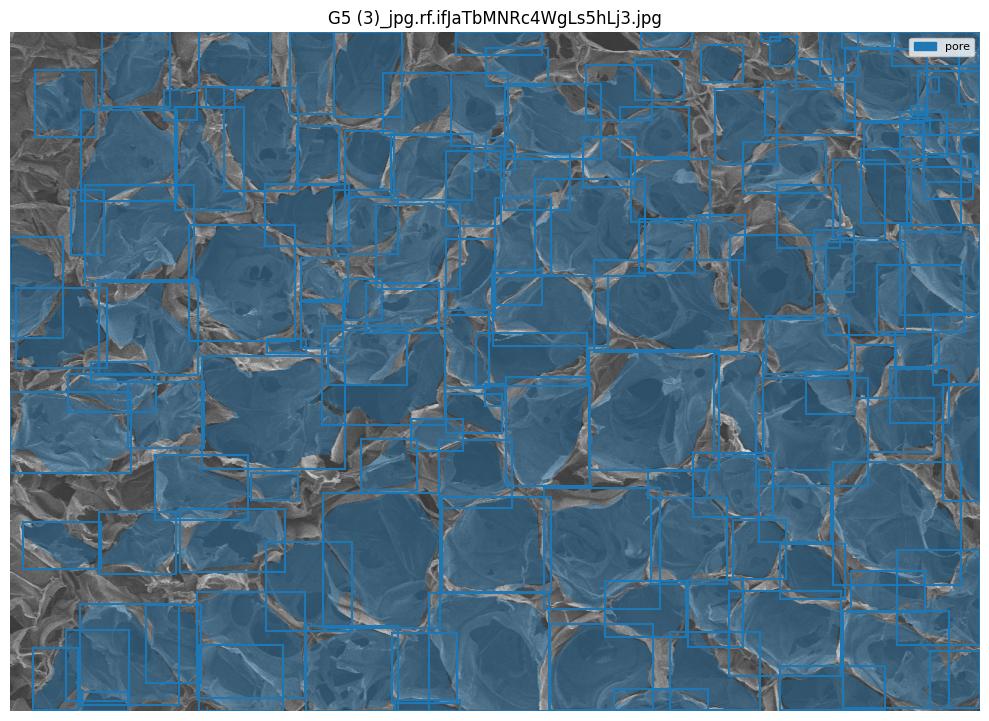} \\
  (m) G5 - Sample 1 & (n) G5 - Sample 2 & (o) G5 - Sample 3
\end{tabular}
\caption{Representative SEM images from the dataset across five magnification series (G1 to G5). Each row shows three selected samples for each G-series group.}
\label{fig:labelled_ausdm_3x5}
\end{figure}

\section{Experiment}

\subsection{Dataset}
The dataset comprises SEM images of tissue-engineered scaffolds fabricated by our group and previously published, cited herein as representative examples \cite{nguyen2025,tran2024}. Briefly, scaffolds were fabricated by blending biopolymers, including alginate, chitosan, gelatin, and silk fibroin, with optional additives, followed by crosslinking and homogeneous mixing. The resulting formulations were frozen and subsequently lyophilized. Samples were then carefully sectioned to expose internal cross-sections, and the resulting microarchitectures were examined by SEM following gold sputter-coating. The scaffold images are documented across five series (G1–G5) of increasing scale, with fields of view ranging from 50µm to 1.00mm, thereby capturing a progressively greater quantity of pores (Fig. \ref{fig:labelled_ausdm_3x5}). Ground-truth labels consisting of bounding box and pixel-level instance segmentation annotations localize individual pore structures. Each sample was manually assessed by domain experts in our group to verify annotation accuracy and to ensure the dataset reflects a realistic deployment scenario.

\subsection{Experiment setup}
All YOLOv26 experiments were conducted on an Apple Silicon M4 
system with 10 GB unified memory, using the Ultralytics 
framework. SAM3 experiments were conducted on a NVIDIA A100 (40 GB) GPU. 
All reported metrics are averaged over three independent runs 
with different random seeds to account for sampling variance 
introduced by the few-shot selection procedure.

\subsection{Evaluation Metrics}
The models are evaluated quantitatively using the following metrics to provide a comparative analysis across three key performance domains: AI-driven detection accuracy, morphological precision, and mechanical regression capabilities:

\begin{enumerate}
    \item \textbf{AI Detection Metrics}:
    \begin{enumerate}
        \item \textbf{IoU} (Intersection over Union): Measures the overlap accuracy between the predicted bounding boxes and the ground-truth pore annotations.
        \item \textbf{Precision, Recall, and F1-Score}: Evaluates the models' robustness in accurately detecting actual pores while minimizing false positives.
    \end{enumerate}
    \item \textbf{Scaffold Morphological Metrics}: Derived directly from the detected bounding boxes, these metrics evaluate how closely the models estimate the physical morphology compared to the ground truth.
    \begin{enumerate}
        \item \textbf{Porosity} ($\phi$): The ratio of the union area of the pore bounding boxes to the total image area:
        $$\phi = \frac{A_{pore}}{A_{image}} * 100$$
        \item \textbf{Material Density} ($\rho_{mat}$): The density of the solid scaffold material:
        $$\rho_{mat} = 1 - \frac{\phi}{100}$$
    \end{enumerate}
    \item \textbf{Mechanical Prediction Metrics}:
    \begin{enumerate}
        \item \textbf{MAE} (Mean Absolute Error): Quantifies the error between the model's predicted mechanical strength and the actual physical testing data
    \end{enumerate}
\end{enumerate}

\section{Results and Discussion}

\subsection{Quantitative Assessment}
Pore detection performance was benchmarked against manual ground truth across the five magnification series (G1 to G5), as summarized in Table~\ref{tab:pore_comparison}. Ground-truth pore counts rise sharply from G1 to G5, from an average of 6.63 pores per image to 129.50, a rise attributable to the larger field of view at lower magnification.

YOLOv26 tracked ground-truth pore counts closely from G1 to G4, returning averages of 8.63, 13.90, 24.54, and 41.36 pores against ground-truth values of 6.63, 16.90, 21.69, and 49.36, respectively. In G5, however, the predicted count plateaued at exactly 60 boxes for both images in the series, irrespective of true pore counts of 116 and 143. Recall thus fell from 0.695 in G1 to 0.345 in G5, while precision remained comparatively stable across the range, between 0.589 and 0.740, indicating that the boxes YOLOv26 did produce were, on the whole, correctly placed.

Averaged across all 44 evaluation images, YOLOv26 detected 25.05 pores per image against a ground truth of 29.68, with a recall of 0.611, precision of 0.633, and F1 score of 0.603. SAM3 detected 28.36 pores per image against the same ground truth, with a recall of 0.573, precision of 0.441, and F1 score of 0.439. YOLOv26 returned a higher precision and a higher overall F1 score, producing three images with predicted porosity below 50\%, compared with eight for SAM3.

\begin{table}[htbp]
\centering
\caption{Comparison of YOLOv26 and SAM3 Models on Pore Prediction}
\label{tab:pore_comparison}
\resizebox{\textwidth}{!}{%
\begin{tabular}{lccccccccc}
\toprule
\textbf{Series} & \textbf{\shortstack{Ground Truth\\Pores}} & \textbf{\shortstack{Predicted Pores\\(YOLO)}} & \textbf{\shortstack{Predicted Pores\\(SAM)}} & \textbf{\shortstack{YOLO\\Recall}} & \textbf{\shortstack{YOLO\\Precision}} & \textbf{\shortstack{YOLO\\F1-Score}} & \textbf{\shortstack{SAM\\Recall}} & \textbf{\shortstack{SAM\\Precision}} & \textbf{\shortstack{SAM\\F1-Score}} \\
\midrule
\textbf{G1} & 6.63   & 8.63  & 15.25 & 0.695          & \textbf{0.589} & \textbf{0.627} & 0.775          & 0.370          & 0.461 \\
\textbf{G2} & 16.90  & 13.90 & 29.40 & 0.607          & \textbf{0.740} & \textbf{0.654} & \textbf{0.649} & 0.349          & 0.446 \\
\textbf{G3} & 21.69  & 24.54 & 28.77 & \textbf{0.612} & \textbf{0.518} & \textbf{0.548} & 0.556          & 0.384          & 0.414 \\
\textbf{G4} & 49.36  & 41.36 & 32.55 & \textbf{0.603} & \textbf{0.685} & \textbf{0.628} & 0.427          & 0.572          & 0.450 \\
\textbf{G5} & 129.50 & 60.00 & 50.00 & \textbf{0.345} & 0.733          & \textbf{0.468} & 0.301          & \textbf{0.822} & 0.427 \\
\bottomrule
\end{tabular}%
}
\end{table}

Porosity mean absolute error (MAE), computed per image and averaged within each series, reveals a crossover between the two models. SAM3 achieved lower porosity error in G1 (15.78\% vs. 23.96\%) and G2 (12.23\% vs. 17.01\%), where pore density is low and its contour-hugging masks cover more of the true void area than YOLOv26's rectangular approximations. The advantage reverses sharply from G3 onward.

\begin{table}[htbp]
\centering
\caption{Comparison of Porosity MAE Between Series YOLO and SAM3 Models}
\label{tab:porosity_mae_comparison}
\begin{tabular}{lcc}
\toprule
\textbf{Series} & \textbf{\shortstack{YOLO Porosity MAE (\%)}} & \textbf{\shortstack{SAM3 Porosity MAE (\%)}} \\
\midrule
\textbf{G1}      & 23.96 & 15.78 \\
\textbf{G2}      & 17.01 & 12.23 \\
\textbf{G3}      & 17.50 & 27.58 \\
\textbf{G4}      & 15.03 & 29.48 \\
\textbf{G5}      & 60.38 & 43.23 \\
\midrule
\textbf{Overall} & 19.90 & 23.13 \\
\bottomrule
\end{tabular}
\end{table}

In G3 and G4, SAM3's porosity MAE more than doubled relative to G2 (27.58\% and 29.48\% against 12.23\%), while YOLOv26's MAE remains below 18\% across the same range. The G5 column is dominated by YOLOv26's detection-ceiling collapse, which elevated its MAE to 60.38\% for those two images, against 43.23\% for SAM3, both unacceptably large. Across the full 44-image evaluation set, YOLOv26 achieved a lower overall porosity MAE of 19.90\%, compared with 23.13\% for SAM3.

\subsection{Qualitative Analysis of Model Predictions}

Table~\ref{tab:scaffold_morphology} shows ground-truth porosity exceeding 86\% in every series, consistent with the highly interconnected, void-dominated architecture typical of biopolymer scaffolds. YOLOv26's predicted porosity tracks this closely through G1 to G4: 72.75\%, 80.53\%, 79.12\%, and 76.19\%, against ground-truth values of 96.39\%, 97.54\%, 96.46\%, and 91.22\%, before collapsing to 28.68\% in G5. The geometry of a rigid bounding box explains the moderate gap in G1 to G4 and the collapse in G5 differently. A rectangular box drawn around an irregular, interconnected pore can either spill into the surrounding polymer matrix, inflating the box beyond the true void, or trim a branching arm of the pore, omitting real void area instead. These two effects roughly cancel through G1 to G4: predicted porosity sits below ground truth, but only moderately. In G5, however, the detection ceiling removes most pores from consideration before this geometric trade-off applies, and predicted porosity thus falls far further than the box-shape mismatch alone would explain.

SAM3's predicted porosity follows a steadier decline, from 84.57\%, 85.30\%, 69.89\%, 61.74\%, to 45.83\% across G1–G5, against ground-truth values remaining within an 89.06–97.54\% band. The decline is non-monotonic: a marginal increase of 0.73 percentage points from G1 to G2, followed by a steep drop of 15.41 points from G2 to G3. Material density, defined as the complement of porosity, mirrors this directly. SAM3's predicted material density rises from 0.1543 in G1 to 0.5417 in G5, a more than 3.5-fold increase, while ground-truth material density rises only from 0.0361 to 0.1094 over the same range. Although SAM3's masks trace pore contours more faithfully than a bounding box at any single magnification, the scaffolds' connective "windows," the narrow polymer bridges separating adjacent pores, allow the model's few-shot segmentation room to merge neighboring voids into a single mask as pore density rises.

\begin{table}[t]
\centering
\caption{Physical Scaffold Morphological Metrics Comparison of YOLOv26 and SAM3}
\label{tab:scaffold_morphology}
\footnotesize
\setlength{\tabcolsep}{5pt}
\begin{tabular}{l c c c c c c}
\toprule
 & \multicolumn{3}{c}{\textbf{Porosity (\%)}} & \multicolumn{3}{c}{\textbf{Material Density}} \\
\cmidrule(lr){2-4} \cmidrule(lr){5-7}
\textbf{Series} & \textbf{GT} & \textbf{SAM} & \textbf{YOLO} & \textbf{GT} & \textbf{SAM} & \textbf{YOLO} \\
\midrule
\textbf{G1} & 96.39 & 84.57 & 72.75 & 0.0361 & 0.1543 & 0.2725 \\
\textbf{G2} & 97.54 & 85.30 & 80.53 & 0.0246 & 0.1470 & 0.1947 \\
\textbf{G3} & 96.46 & 69.89 & 79.12 & 0.0354 & 0.3011 & 0.2088 \\
\textbf{G4} & 91.22 & 61.74 & 76.19 & 0.0878 & 0.3826 & 0.2381 \\
\textbf{G5} & 89.06 & 45.83 & 28.68 & 0.1094 & 0.5417 & 0.7132 \\
\bottomrule
\end{tabular}
\end{table}

\begin{figure}[htbp]
\centering
\includegraphics[width=\textwidth]{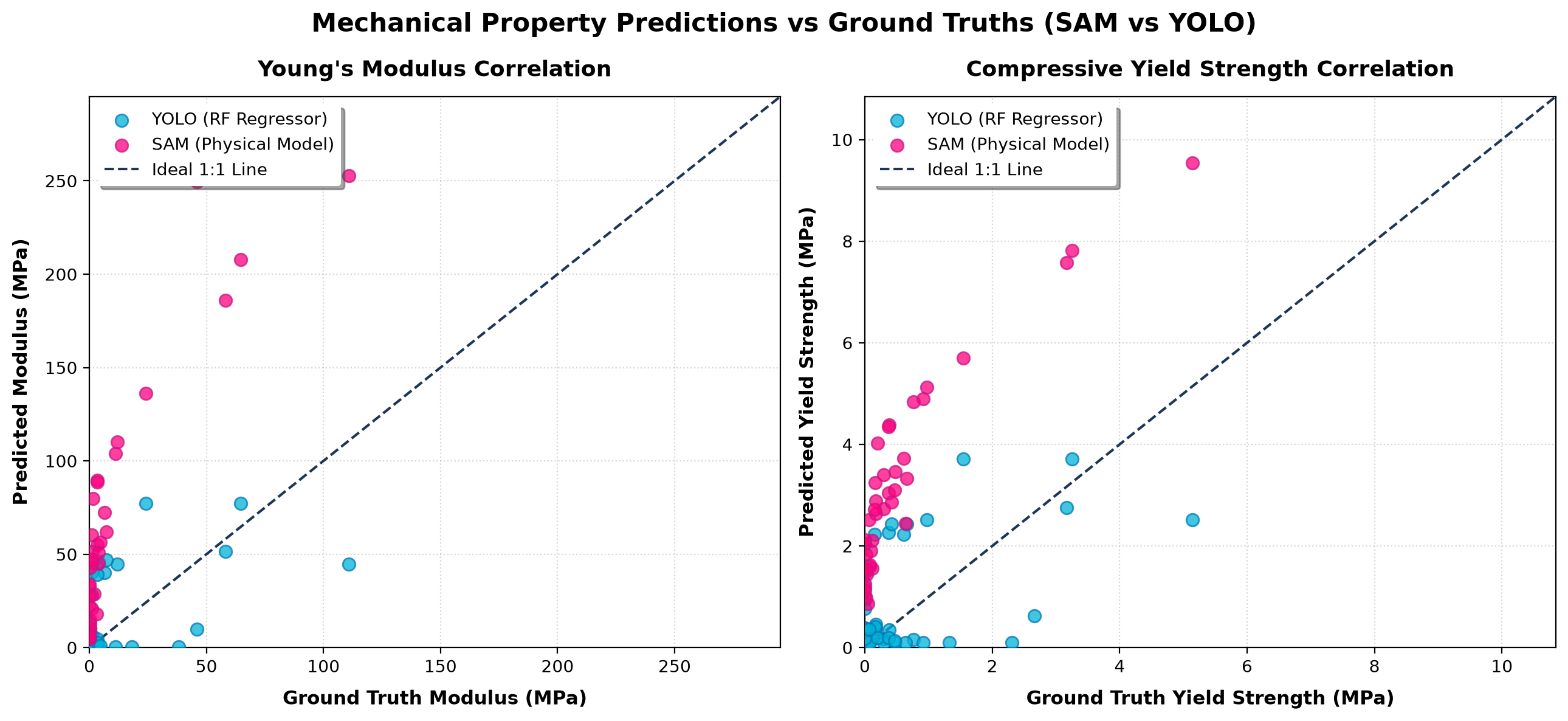}
\caption{Mechanical property prediction correlation plots for Young's Modulus (left) and Compressive Yield Strength (right) comparing YOLOv26s (RF Regressor) and SAM3 (Gibson-Ashby Physical Model) against manual ground truth measurements.}
\label{fig:mechanical_correlation}
\end{figure}

Figure ~\ref{fig:mechanical_correlation} extends this comparison from morphology to mechanical-property prediction. Neither regressor tracks the 1:1 reference line closely across its full range. YOLOv26's Random Forest regressor clusters its Young's modulus predictions within a comparatively narrow band, mostly under 100 MPa, even where ground-truth modulus exceeds 250 MPa, indicating systematic under-prediction of stiffness for the stiffest scaffolds in the test set. SAM3's Gibson-Ashby physical model shows the opposite tendency for compressive yield strength: several specimens with a ground-truth yield strength below 2 MPa receive predicted values between 6 and 10 MPa. This over-prediction is consistent with the gradual porosity under-estimation in Table ~\ref{tab:scaffold_morphology}, since the Gibson-Ashby relation predicts higher strength as porosity decreases for a given material. Both regressors inherit, in different directions, the morphological error already present at the pore-detection stage. YOLOv26's narrower predicted range likely reflects the limited diversity of porosity values its bounding boxes, while SAM3's high-end over-predictions concentrate at the same low-porosity, high-density scaffolds where Table ~\ref{tab:scaffold_morphology} shows its largest morphological error.

\subsection{Failure Case Analysis}


Catastrophic failure, defined here as predicted porosity (Table~\ref{tab:scaffold_morphology}) falling below 50\% of true scaffold void area, occurs in three YOLOv26 images and eight SAM3 images. The two failure sets concentrate in different regions of the dataset and arise from different mechanisms.

\begin{figure}[htbp]
\centering
\setlength{\tabcolsep}{4pt}
\begin{tabular}{ccc}
  \includegraphics[width=0.30\textwidth]{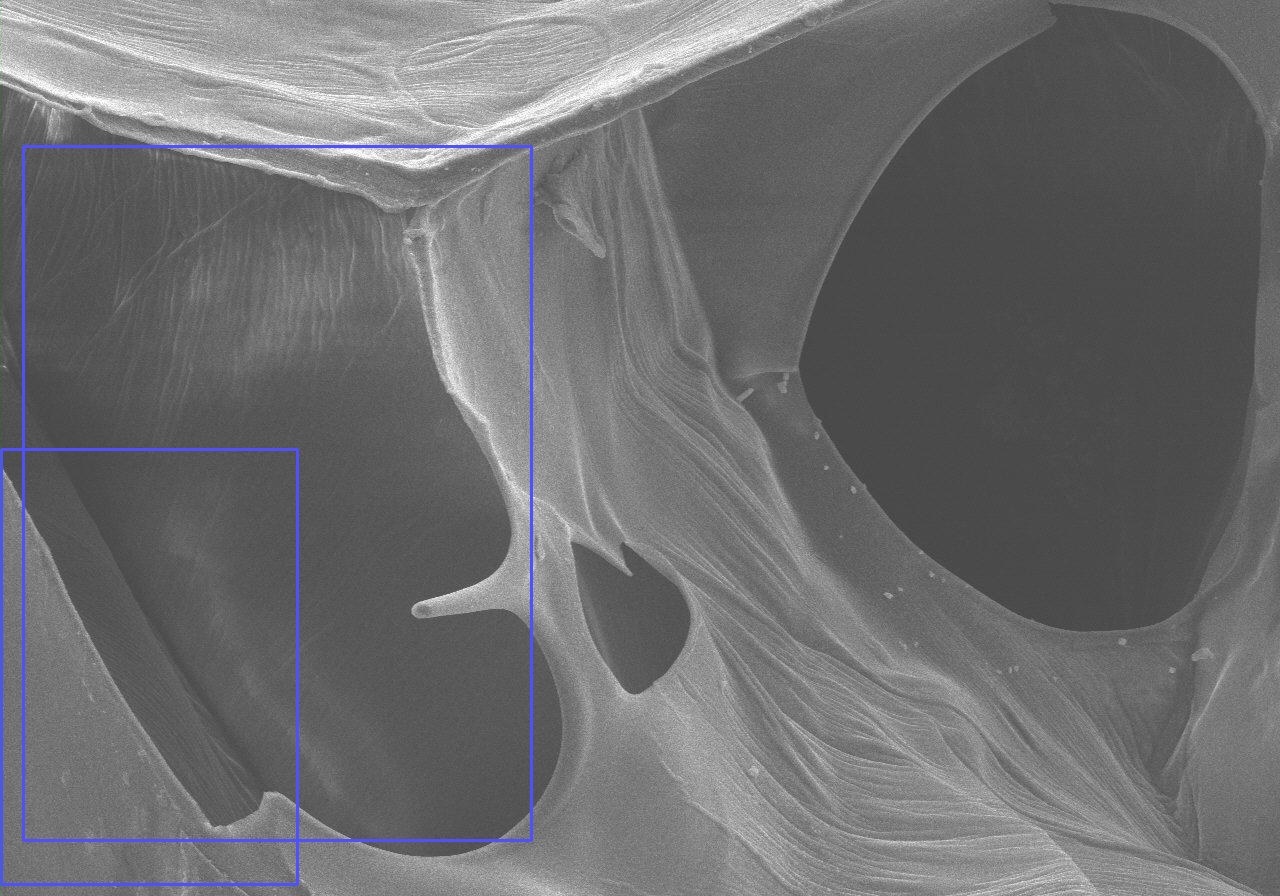} &
  \includegraphics[width=0.30\textwidth]{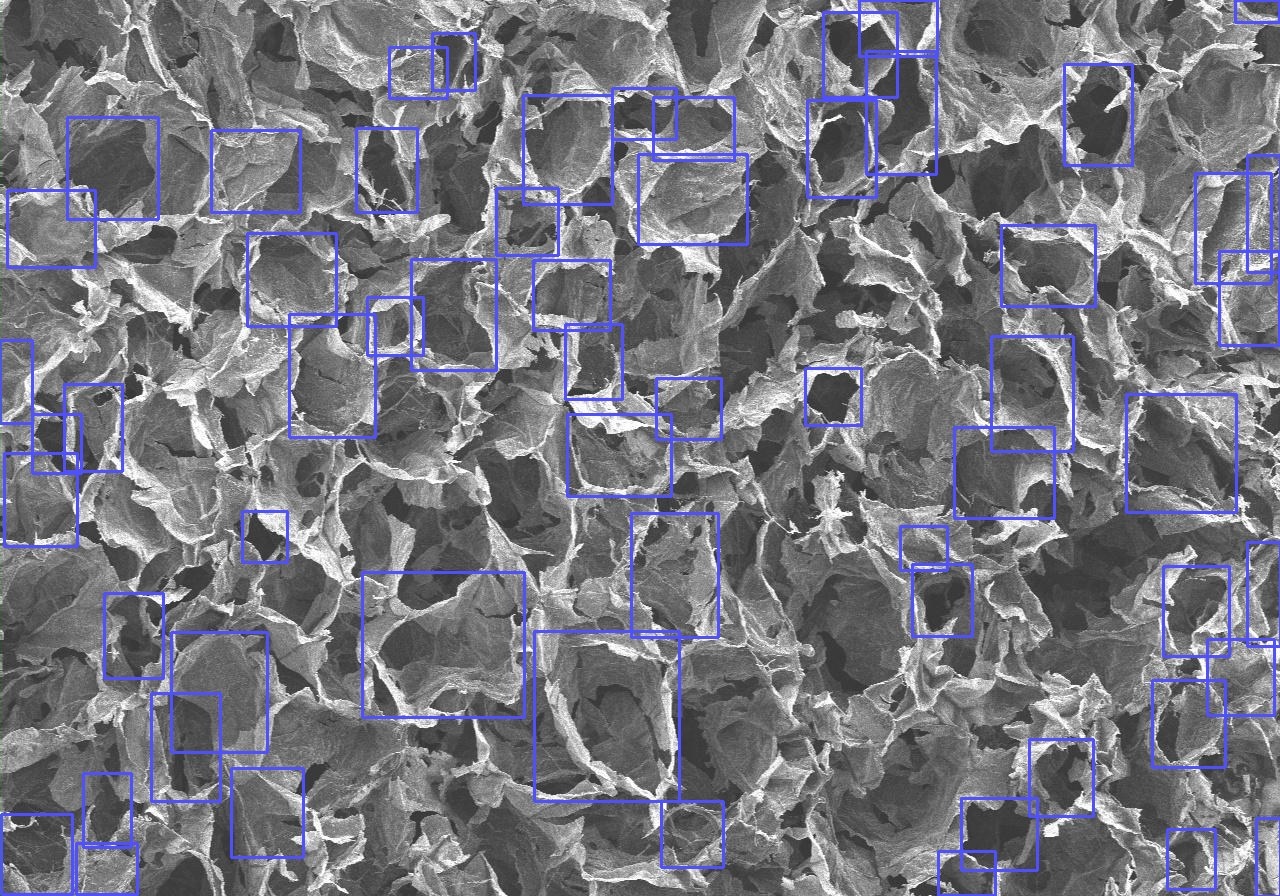} &
  \includegraphics[width=0.30\textwidth]{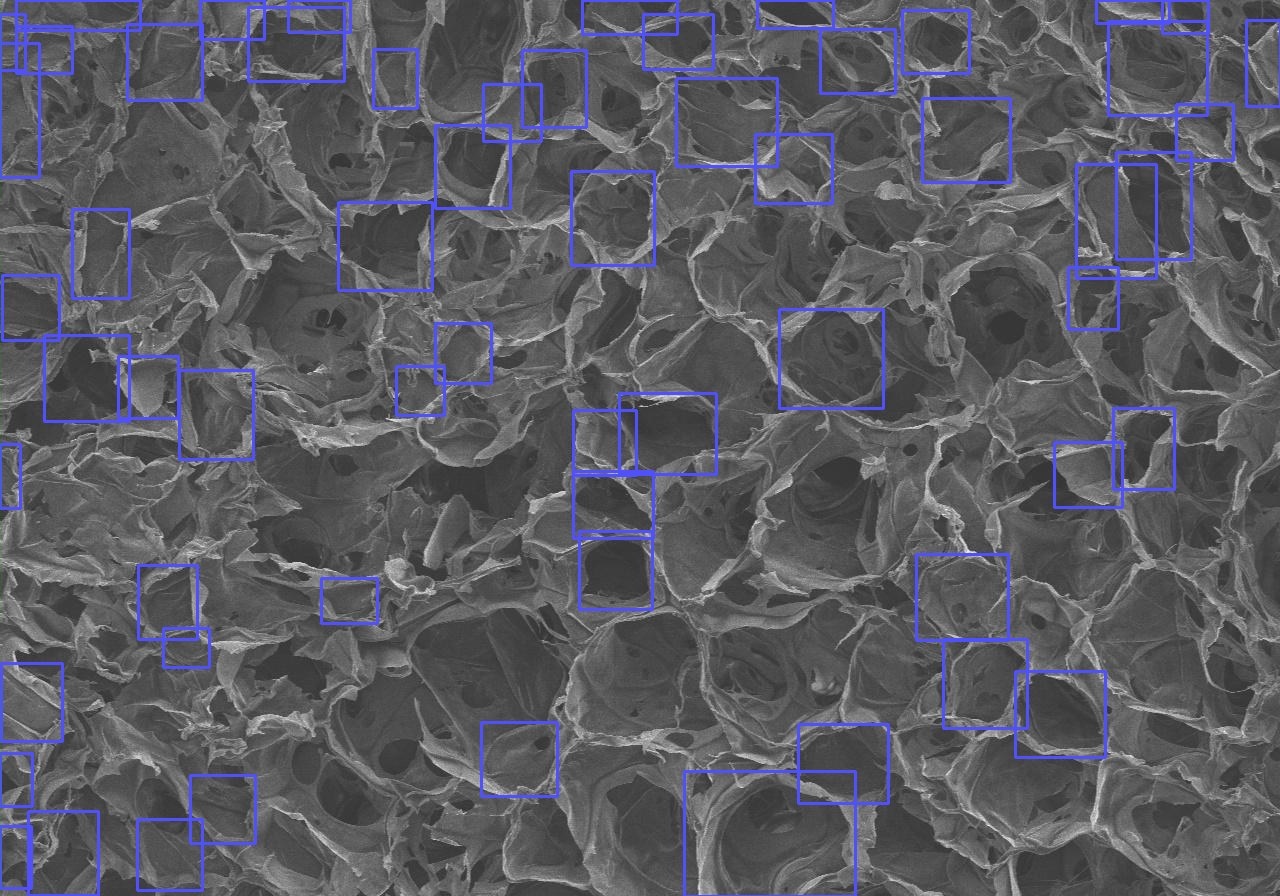} \\
  (a) G1 (6) - Pred: 32.78\% & (b) G5 (2) - Pred: 32.22\% & (c) G5 (3) - Pred: 25.14\% \\
\end{tabular}
\caption{YOLO detection failure cases where the predicted porosity is under 50\%.}
\label{fig:yolo_failures}
\end{figure}

YOLOv26's failures consist of two images from the G5 series (Fig.~\ref{fig:yolo_failures}b-c) and one case in G1 (Fig.~\ref{fig:yolo_failures}a). Both G5 images return exactly 60 predicted boxes, regardless of true pore counts of 116 and 143, respectively. The model does not fail to detect the additional pores so much as discard them once a fixed output ceiling is reached. The remained boxes still align with real pores (G5 series precision is 0.733) but the 56 to 83 missing boxes per image collapse predicted porosity to 32.22\% and 25.14\%, against ground-truth values of 91.70\% and 86.42\%. The single G1 failure, G1(6), follows a different mechanism. Its ground-truth annotation comprises a single near-frame-filling void with two small secondary pores. YOLOv26's box for the dominant void captures only a small fraction of its true extent, driving material density from 0.0008 in ground truth to 0.672 in the prediction, the largest single-image discrepancy for either model in the dataset.

\begin{figure}[htbp]
\centering
\setlength{\tabcolsep}{4pt}
\begin{tabular}{ccc}
  \includegraphics[width=0.30\textwidth]{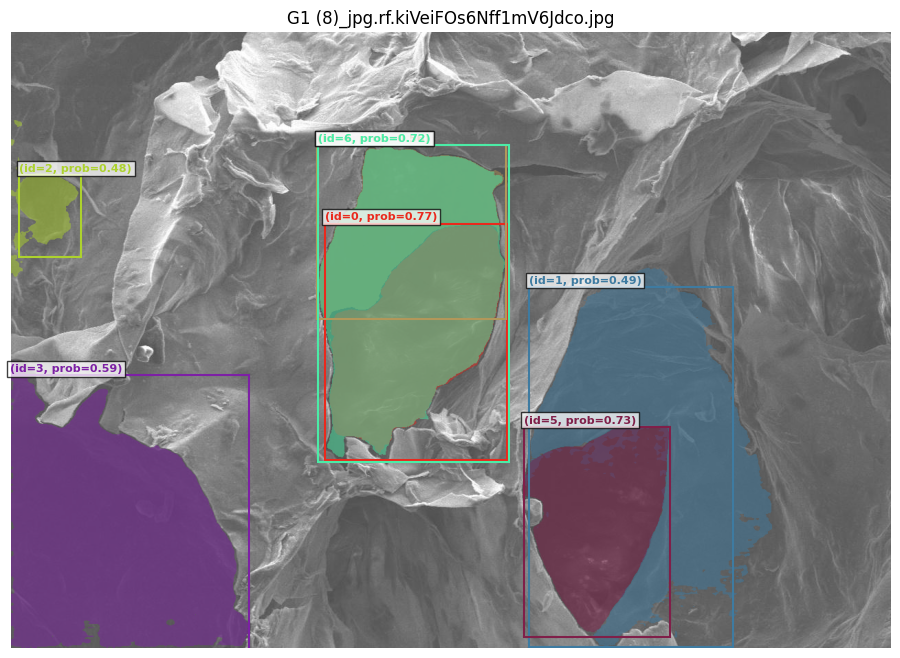} &
  \includegraphics[width=0.30\textwidth]{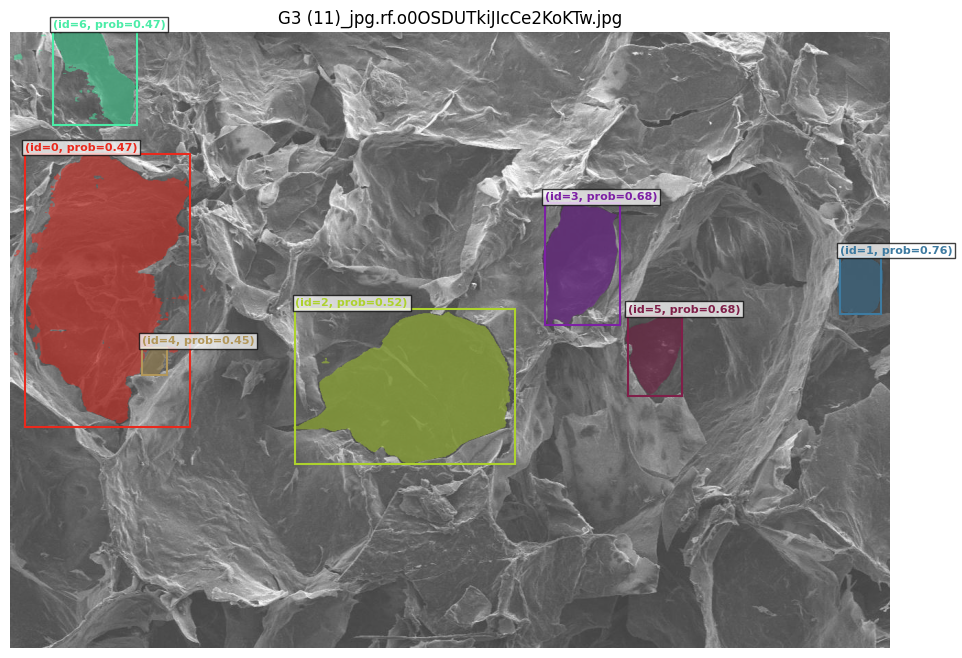} &
  \includegraphics[width=0.30\textwidth]{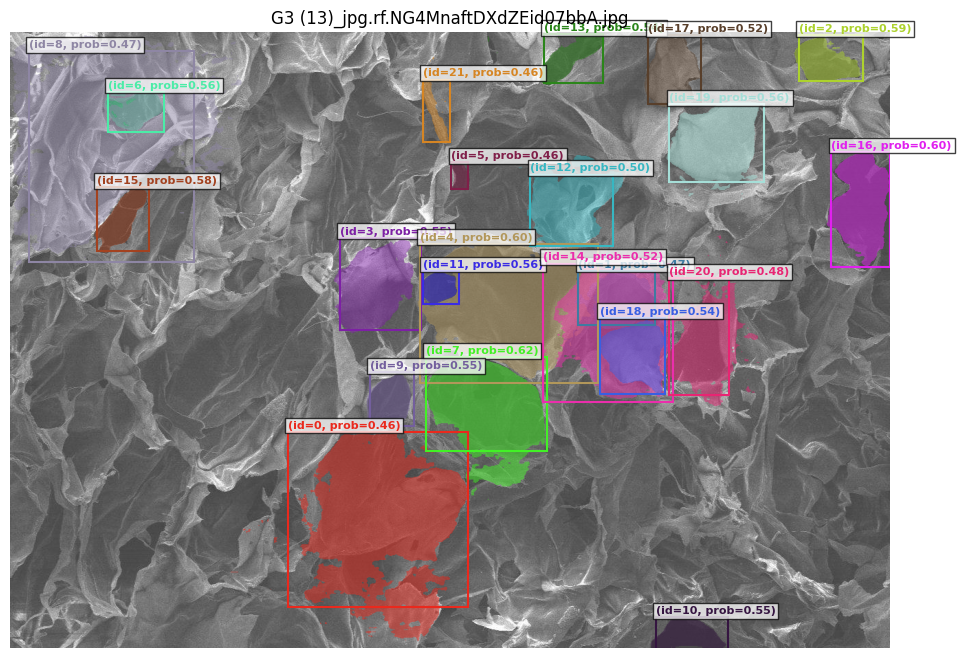} \\
  (a) G1 (8) - Pred: 35.53\% & (b) G3 (11) - Pred: 22.27\% & (c) G3 (13) - Pred: 32.26\% \\[6pt]

  \includegraphics[width=0.30\textwidth]{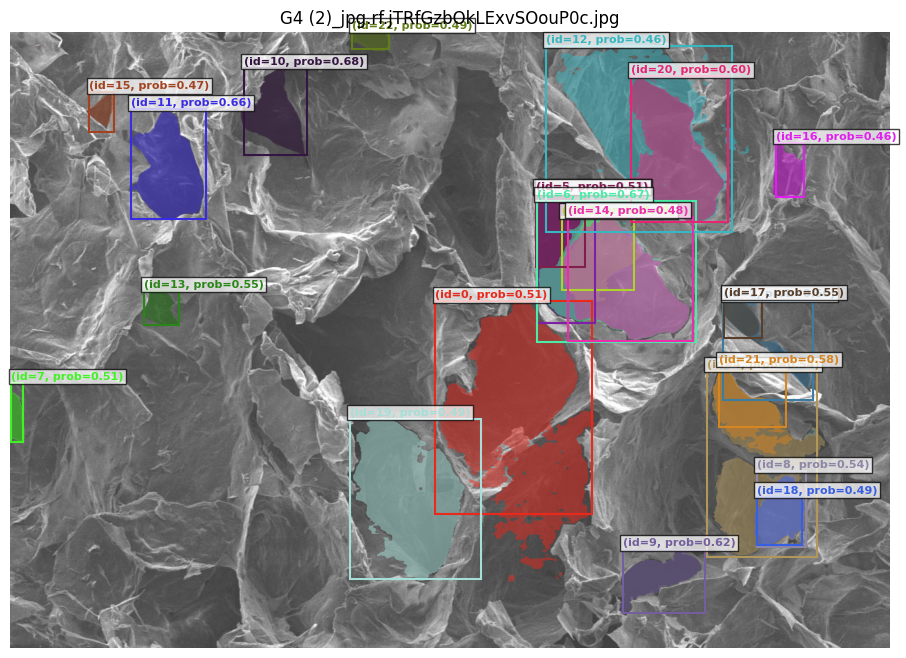} &
  \includegraphics[width=0.30\textwidth]{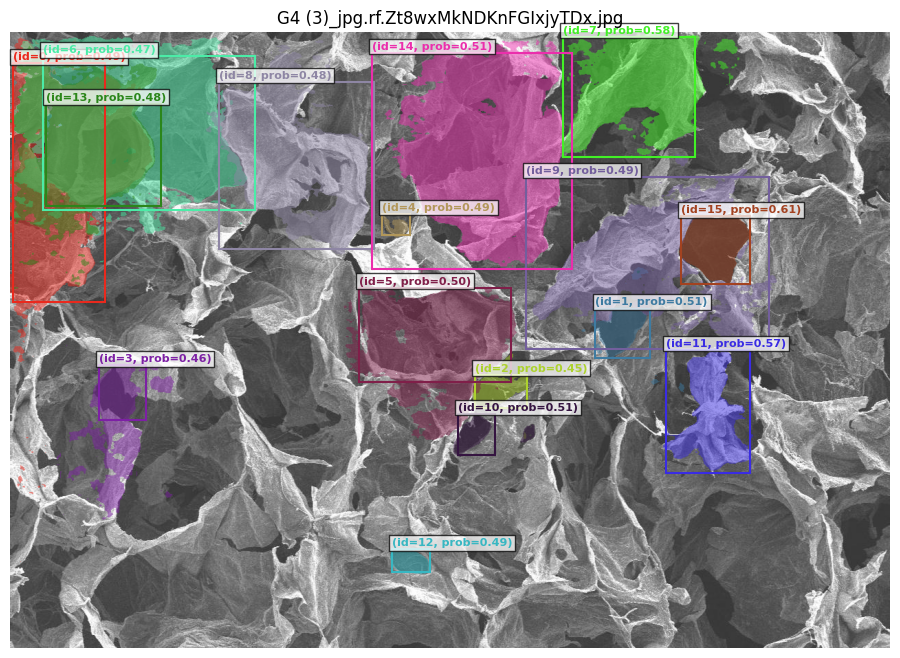} &
  \includegraphics[width=0.30\textwidth]{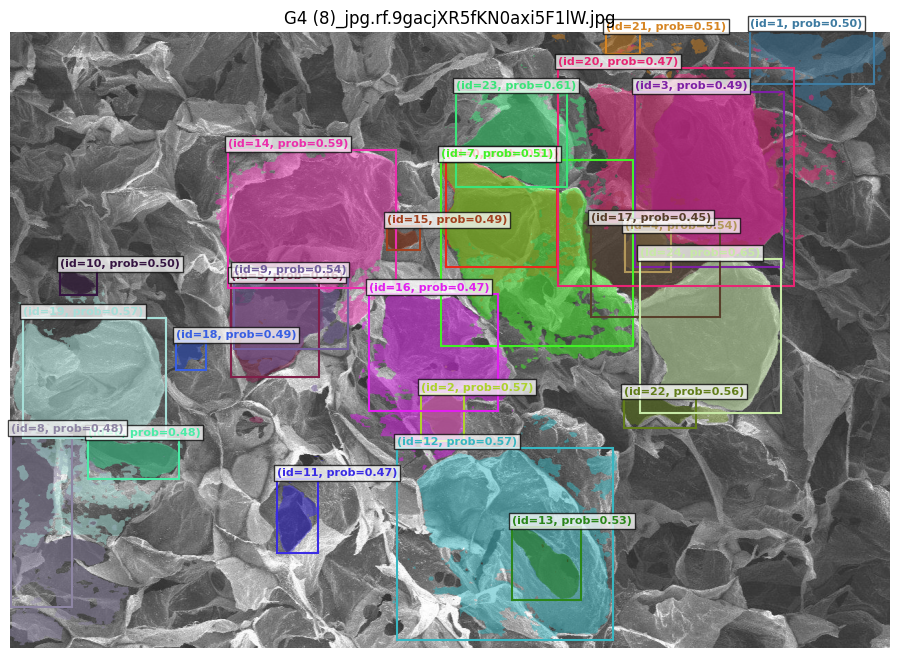} \\
  (d) G4 (2) - Pred: 31.10\% & (e) G4 (3) - Pred: 41.98\% & (f) G4 (8) - Pred: 46.38\% \\[6pt]

  \includegraphics[width=0.30\textwidth]{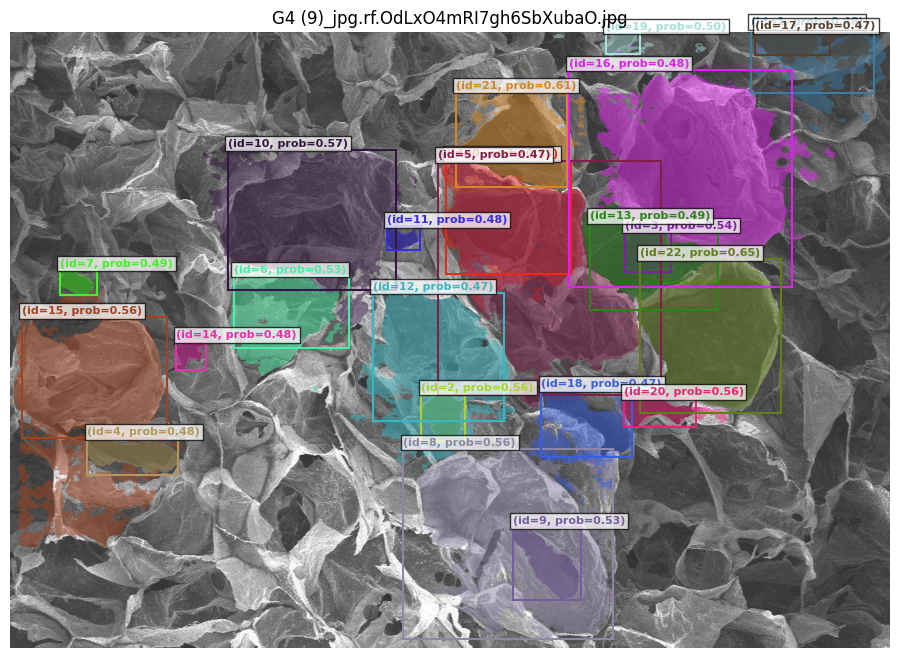} &
  \includegraphics[width=0.30\textwidth]{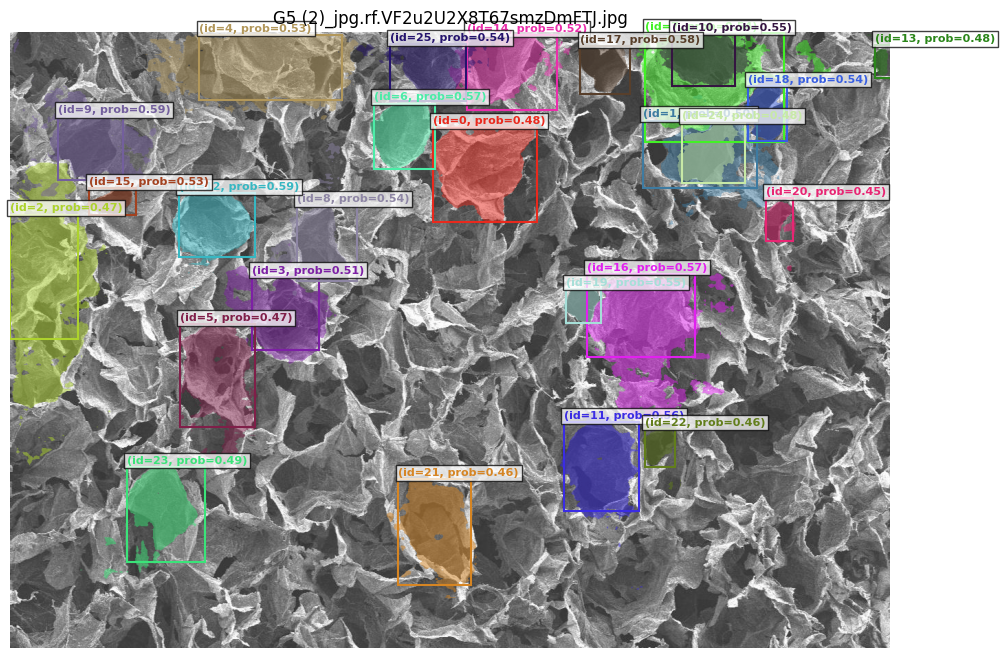} \\
  (g) G4 (9) - Pred: 46.07\%  & (h) G5 (2) - Pred: 26.95\%
\end{tabular}
\caption{SAM3 segmentation failure cases where the predicted porosity is under 50\%.}
\label{fig:sam_failures}
\end{figure}

SAM3's failures (Fig.~\ref{fig:sam_failures}a-h) sit predominantly in the three densest series, G3, G4, and G5, with one exception in G1. Predicted porosity for the nine failing images ranges from 22.27\% (G3(11)) to 46.38\% (G4(8)), against ground-truth values that remain above 83\% in every case. Two effects appear together in the segmentation overlays as density rises. First, adjacent pores separated only by thin polymer "windows" get absorbed into single, larger masks: G3(11) returns 7 predicted pores against a ground truth of 30. Second, entire regions of true void area receive no mask at all in the most crowded frames. The second effect carries more weight than the first. Pure consolidation of existing detections into fewer, larger masks would lower pore count without lowering covered area by much, yet porosity in the densest failures drops by as much as 75.27 percentage points in G3(11), from 97.54\% to 22.27\%, and 64.75 points in G5(2), from 91.70\% to 26.95\%, far more than count consolidation alone would produce. Mask coverage, not just mask count, therefore breaks down once the scene becomes sufficiently crowded. Among the four G4 failures, G4(2) and G4(3) are the most severe, with predicted porosity of 31.10\% and 41.98\% respectively. G4(8) and G4(9) sit just below threshold at 46.38\% and 46.07\%. All four share the same failure pattern: very low recall (0.19 to 0.22) against comparatively high precision (0.63 to 0.70), meaning the masks that SAM3 does produce are geometrically correct but coverage of the full pore field is grossly incomplete. The single exception, G1(8), occurs at much lower density, with only 8 pores in ground truth, yet predicted porosity still collapses to 35.53\%. The overlay for this image shows several pores receiving no mask at all despite the comparatively sparse, well-separated field. This points to a localized contrast or illumination artefact specific to this frame rather than the crowding effect mentioned above.

Across both models, the failures separate cleanly along one axis: the number of discrete objects the input image actually contains. YOLOv26's box-based architecture fails once this count exceeds its fixed output limit. While SAM3's mask-based architecture fails once nearby objects become close enough, and crowded enough, for mask coverage itself to break down. Image magnification by itself does not predict failure for either model: G1, the highest-magnification series in the dataset, contributes one failure for each model, while the majority of both failure sets resides at the opposite end of the scale, in the densest, lowest-magnification images.

\section{Conclusion}

Pore detection and morphological characterization were benchmarked across two foundation models, fine-tuned with only 12 SEM images of biopolymer bone scaffolds, against manual ground truth spanning five magnification series. The two architectures achieve broadly similar overall accuracy by different routes, and they fail under different, identifiable conditions.

YOLOv26 tracks ground-truth pore counts and porosity closely at low-to-moderate pore density and breaks down only once the number of true pores in a single frame exceeds its detection budget. Beyond this point, predicted porosity collapses regardless of how well-separated or well-defined the missed pores were. SAM3 over-detects pores at low density and under-detects at high density, and unlike YOLOv26, its porosity error degrades gradually rather than abruptly, from roughly 15.78 percentage points at the sparsest end of the dataset (G1) to almost 43.23 points at the densest end (G5). Two distinct effects drive SAM3's degradation at high density: mask consolidation across thin polymer "windows" lowers pore count without substantial loss of covered area, while a separate failure of mask coverage in the most crowded scenes drives most of SAM3's worst-case porosity error. Averaged across the full 56-image dataset, YOLOv26 achieves a lower porosity error, higher precision, and a higher F1 score than SAM3, while SAM3 retains the advantage of pixel-accurate contours that a bounding box cannot reproduce.

These findings point towards a density-aware rather than magnification-aware deployment strategy. A bounding-box detector is the more reliable default at low-to-moderate pore counts, provided its detection ceiling is set high enough, or its inference is tiled, for the densest frames it will encounter. A mask-based segmenter remains preferable wherever contour-level shape matters more than aggregate area accuracy, but its few-shot performance should be expected to degrade as scaffold pore density increases. That degradation should be checked directly rather than assumed to be a marginal effect.

Several future research directions emerge from this work. Building on our central finding that failure tracks pore density rather than magnification, a natural next step is a density-aware pipeline that routes each image to the better-suited detector, or a hybrid model that uses YOLOv26's robust bounding-box detections as prompts to guide SAM3's contour-precise segmentation. The mechanical-property prediction stage also merits further investigation, specifically the replacement of the current regressors with physics-informed models validated against a larger body of destructive-test measurements. Finally, the framework could be extended to additional scaffold formulations, fabrication batches, and imaging modalities such as micro-CT, while expanding the expert-annotated dataset that few-shot methods in this domain critically depend on.

\begin{credits}
\subsubsection*{Disclosure of Interests.}
The authors have no competing interests to declare that are relevant to the content of this article.
\end{credits}

%
%
%
%

\end{document}